\documentclass[letterpaper, 10 pt, conference]{ieeeconf}
\IEEEoverridecommandlockouts
\usepackage{pstricks}
\usepackage{ifplatform}
\usepackage{xkeyval}
\usepackage{rotating}

\usepackage{graphics} 
\usepackage{times} 
\usepackage{amsmath} 
\usepackage{psfrag}
\usepackage{cancel}

\usepackage[normalem]{ulem}
\usepackage{cite}
\usepackage{url}
\usepackage{color}
\definecolor{AliceBlue}{rgb}{0.94,0.97,1.00}
\definecolor{AntiqueWhite1}{rgb}{1.00,0.94,0.86}
\definecolor{AntiqueWhite2}{rgb}{0.93,0.87,0.80}
\definecolor{AntiqueWhite3}{rgb}{0.80,0.75,0.69}
\definecolor{AntiqueWhite4}{rgb}{0.55,0.51,0.47}
\definecolor{AntiqueWhite}{rgb}{0.98,0.92,0.84}
\definecolor{BlanchedAlmond}{rgb}{1.00,0.92,0.80}
\definecolor{BlueViolet}{rgb}{0.54,0.17,0.89}
\definecolor{CadetBlue1}{rgb}{0.60,0.96,1.00}
\definecolor{CadetBlue2}{rgb}{0.56,0.90,0.93}
\definecolor{CadetBlue3}{rgb}{0.48,0.77,0.80}
\definecolor{CadetBlue4}{rgb}{0.33,0.53,0.55}
\definecolor{CadetBlue}{rgb}{0.37,0.62,0.63}
\definecolor{CornflowerBlue}{rgb}{0.39,0.58,0.93}
\definecolor{DarkBlue}{rgb}{0.00,0.00,0.55}
\definecolor{DarkCyan}{rgb}{0.00,0.55,0.55}
\definecolor{DarkGoldenrod1}{rgb}{1.00,0.73,0.06}
\definecolor{DarkGoldenrod2}{rgb}{0.93,0.68,0.05}
\definecolor{DarkGoldenrod3}{rgb}{0.80,0.58,0.05}
\definecolor{DarkGoldenrod4}{rgb}{0.55,0.40,0.03}
\definecolor{DarkGoldenrod}{rgb}{0.72,0.53,0.04}
\definecolor{DarkGray}{rgb}{0.66,0.66,0.66}
\definecolor{DarkGreen}{rgb}{0.00,0.39,0.00}
\definecolor{DarkGrey}{rgb}{0.66,0.66,0.66}
\definecolor{DarkKhaki}{rgb}{0.74,0.72,0.42}
\definecolor{DarkMagenta}{rgb}{0.55,0.00,0.55}
\definecolor{DarkOliveGreen1}{rgb}{0.79,1.00,0.44}
\definecolor{DarkOliveGreen2}{rgb}{0.74,0.93,0.41}
\definecolor{DarkOliveGreen3}{rgb}{0.64,0.80,0.35}
\definecolor{DarkOliveGreen4}{rgb}{0.43,0.55,0.24}
\definecolor{DarkOliveGreen}{rgb}{0.33,0.42,0.18}
\definecolor{DarkOrange1}{rgb}{1.00,0.50,0.00}
\definecolor{DarkOrange2}{rgb}{0.93,0.46,0.00}
\definecolor{DarkOrange3}{rgb}{0.80,0.40,0.00}
\definecolor{DarkOrange4}{rgb}{0.55,0.27,0.00}
\definecolor{DarkOrange}{rgb}{1.00,0.55,0.00}
\definecolor{DarkOrchid1}{rgb}{0.75,0.24,1.00}
\definecolor{DarkOrchid2}{rgb}{0.70,0.23,0.93}
\definecolor{DarkOrchid3}{rgb}{0.60,0.20,0.80}
\definecolor{DarkOrchid4}{rgb}{0.41,0.13,0.55}
\definecolor{DarkOrchid}{rgb}{0.60,0.20,0.80}
\definecolor{DarkRed}{rgb}{0.55,0.00,0.00}
\definecolor{DarkSalmon}{rgb}{0.91,0.59,0.48}
\definecolor{DarkSeaGreen1}{rgb}{0.76,1.00,0.76}
\definecolor{DarkSeaGreen2}{rgb}{0.71,0.93,0.71}
\definecolor{DarkSeaGreen3}{rgb}{0.61,0.80,0.61}
\definecolor{DarkSeaGreen4}{rgb}{0.41,0.55,0.41}
\definecolor{DarkSeaGreen}{rgb}{0.56,0.74,0.56}
\definecolor{DarkSlateBlue}{rgb}{0.28,0.24,0.55}
\definecolor{DarkSlateGray1}{rgb}{0.59,1.00,1.00}
\definecolor{DarkSlateGray2}{rgb}{0.55,0.93,0.93}
\definecolor{DarkSlateGray3}{rgb}{0.47,0.80,0.80}
\definecolor{DarkSlateGray4}{rgb}{0.32,0.55,0.55}
\definecolor{DarkSlateGray}{rgb}{0.18,0.31,0.31}
\definecolor{DarkSlateGrey}{rgb}{0.18,0.31,0.31}
\definecolor{DarkTurquoise}{rgb}{0.00,0.81,0.82}
\definecolor{DarkViolet}{rgb}{0.58,0.00,0.83}
\definecolor{DeepPink1}{rgb}{1.00,0.08,0.58}
\definecolor{DeepPink2}{rgb}{0.93,0.07,0.54}
\definecolor{DeepPink3}{rgb}{0.80,0.06,0.46}
\definecolor{DeepPink4}{rgb}{0.55,0.04,0.31}
\definecolor{DeepPink}{rgb}{1.00,0.08,0.58}
\definecolor{DeepSkyBlue1}{rgb}{0.00,0.75,1.00}
\definecolor{DeepSkyBlue2}{rgb}{0.00,0.70,0.93}
\definecolor{DeepSkyBlue3}{rgb}{0.00,0.60,0.80}
\definecolor{DeepSkyBlue4}{rgb}{0.00,0.41,0.55}
\definecolor{DeepSkyBlue}{rgb}{0.00,0.75,1.00}
\definecolor{DimGray}{rgb}{0.41,0.41,0.41}
\definecolor{DimGrey}{rgb}{0.41,0.41,0.41}
\definecolor{DodgerBlue1}{rgb}{0.12,0.56,1.00}
\definecolor{DodgerBlue2}{rgb}{0.11,0.53,0.93}
\definecolor{DodgerBlue3}{rgb}{0.09,0.45,0.80}
\definecolor{DodgerBlue4}{rgb}{0.06,0.31,0.55}
\definecolor{DodgerBlue}{rgb}{0.12,0.56,1.00}
\definecolor{FloralWhite}{rgb}{1.00,0.98,0.94}
\definecolor{ForestGreen}{rgb}{0.13,0.55,0.13}
\definecolor{GhostWhite}{rgb}{0.97,0.97,1.00}
\definecolor{GreenYellow}{rgb}{0.68,1.00,0.18}
\definecolor{HotPink1}{rgb}{1.00,0.43,0.71}
\definecolor{HotPink2}{rgb}{0.93,0.42,0.65}
\definecolor{HotPink3}{rgb}{0.80,0.38,0.56}
\definecolor{HotPink4}{rgb}{0.55,0.23,0.38}
\definecolor{HotPink}{rgb}{1.00,0.41,0.71}
\definecolor{IndianRed1}{rgb}{1.00,0.42,0.42}
\definecolor{IndianRed2}{rgb}{0.93,0.39,0.39}
\definecolor{IndianRed3}{rgb}{0.80,0.33,0.33}
\definecolor{IndianRed4}{rgb}{0.55,0.23,0.23}
\definecolor{IndianRed}{rgb}{0.80,0.36,0.36}
\definecolor{LavenderBlush1}{rgb}{1.00,0.94,0.96}
\definecolor{LavenderBlush2}{rgb}{0.93,0.88,0.90}
\definecolor{LavenderBlush3}{rgb}{0.80,0.76,0.77}
\definecolor{LavenderBlush4}{rgb}{0.55,0.51,0.53}
\definecolor{LavenderBlush}{rgb}{1.00,0.94,0.96}
\definecolor{LawnGreen}{rgb}{0.49,0.99,0.00}
\definecolor{LemonChiffon1}{rgb}{1.00,0.98,0.80}
\definecolor{LemonChiffon2}{rgb}{0.93,0.91,0.75}
\definecolor{LemonChiffon3}{rgb}{0.80,0.79,0.65}
\definecolor{LemonChiffon4}{rgb}{0.55,0.54,0.44}
\definecolor{LemonChiffon}{rgb}{1.00,0.98,0.80}
\definecolor{LightBlue1}{rgb}{0.75,0.94,1.00}
\definecolor{LightBlue2}{rgb}{0.70,0.87,0.93}
\definecolor{LightBlue3}{rgb}{0.60,0.75,0.80}
\definecolor{LightBlue4}{rgb}{0.41,0.51,0.55}
\definecolor{LightBlue}{rgb}{0.68,0.85,0.90}
\definecolor{LightCoral}{rgb}{0.94,0.50,0.50}
\definecolor{LightCyan1}{rgb}{0.88,1.00,1.00}
\definecolor{LightCyan2}{rgb}{0.82,0.93,0.93}
\definecolor{LightCyan3}{rgb}{0.71,0.80,0.80}
\definecolor{LightCyan4}{rgb}{0.48,0.55,0.55}
\definecolor{LightCyan}{rgb}{0.88,1.00,1.00}
\definecolor{LightGoldenrod1}{rgb}{1.00,0.93,0.55}
\definecolor{LightGoldenrod2}{rgb}{0.93,0.86,0.51}
\definecolor{LightGoldenrod3}{rgb}{0.80,0.75,0.44}
\definecolor{LightGoldenrod4}{rgb}{0.55,0.51,0.30}
\definecolor{LightGoldenrodYellow}{rgb}{0.98,0.98,0.82}
\definecolor{LightGoldenrod}{rgb}{0.93,0.87,0.51}
\definecolor{LightGray}{rgb}{0.83,0.83,0.83}
\definecolor{LightGreen}{rgb}{0.56,0.93,0.56}
\definecolor{LightGrey}{rgb}{0.83,0.83,0.83}
\definecolor{LightPink1}{rgb}{1.00,0.68,0.73}
\definecolor{LightPink2}{rgb}{0.93,0.64,0.68}
\definecolor{LightPink3}{rgb}{0.80,0.55,0.58}
\definecolor{LightPink4}{rgb}{0.55,0.37,0.40}
\definecolor{LightPink}{rgb}{1.00,0.71,0.76}
\definecolor{LightSalmon1}{rgb}{1.00,0.63,0.48}
\definecolor{LightSalmon2}{rgb}{0.93,0.58,0.45}
\definecolor{LightSalmon3}{rgb}{0.80,0.51,0.38}
\definecolor{LightSalmon4}{rgb}{0.55,0.34,0.26}
\definecolor{LightSalmon}{rgb}{1.00,0.63,0.48}
\definecolor{LightSeaGreen}{rgb}{0.13,0.70,0.67}
\definecolor{LightSkyBlue1}{rgb}{0.69,0.89,1.00}
\definecolor{LightSkyBlue2}{rgb}{0.64,0.83,0.93}
\definecolor{LightSkyBlue3}{rgb}{0.55,0.71,0.80}
\definecolor{LightSkyBlue4}{rgb}{0.38,0.48,0.55}
\definecolor{LightSkyBlue}{rgb}{0.53,0.81,0.98}
\definecolor{LightSlateBlue}{rgb}{0.52,0.44,1.00}
\definecolor{LightSlateGray}{rgb}{0.47,0.53,0.60}
\definecolor{LightSlateGrey}{rgb}{0.47,0.53,0.60}
\definecolor{LightSteelBlue1}{rgb}{0.79,0.88,1.00}
\definecolor{LightSteelBlue2}{rgb}{0.74,0.82,0.93}
\definecolor{LightSteelBlue3}{rgb}{0.64,0.71,0.80}
\definecolor{LightSteelBlue4}{rgb}{0.43,0.48,0.55}
\definecolor{LightSteelBlue}{rgb}{0.69,0.77,0.87}
\definecolor{LightYellow1}{rgb}{1.00,1.00,0.88}
\definecolor{LightYellow2}{rgb}{0.93,0.93,0.82}
\definecolor{LightYellow3}{rgb}{0.80,0.80,0.71}
\definecolor{LightYellow4}{rgb}{0.55,0.55,0.48}
\definecolor{LightYellow}{rgb}{1.00,1.00,0.88}
\definecolor{LimeGreen}{rgb}{0.20,0.80,0.20}
\definecolor{MediumAquamarine}{rgb}{0.40,0.80,0.67}
\definecolor{MediumBlue}{rgb}{0.00,0.00,0.80}
\definecolor{MediumOrchid1}{rgb}{0.88,0.40,1.00}
\definecolor{MediumOrchid2}{rgb}{0.82,0.37,0.93}
\definecolor{MediumOrchid3}{rgb}{0.71,0.32,0.80}
\definecolor{MediumOrchid4}{rgb}{0.48,0.22,0.55}
\definecolor{MediumOrchid}{rgb}{0.73,0.33,0.83}
\definecolor{MediumPurple1}{rgb}{0.67,0.51,1.00}
\definecolor{MediumPurple2}{rgb}{0.62,0.47,0.93}
\definecolor{MediumPurple3}{rgb}{0.54,0.41,0.80}
\definecolor{MediumPurple4}{rgb}{0.36,0.28,0.55}
\definecolor{MediumPurple}{rgb}{0.58,0.44,0.86}
\definecolor{MediumSeaGreen}{rgb}{0.24,0.70,0.44}
\definecolor{MediumSlateBlue}{rgb}{0.48,0.41,0.93}
\definecolor{MediumSpringGreen}{rgb}{0.00,0.98,0.60}
\definecolor{MediumTurquoise}{rgb}{0.28,0.82,0.80}
\definecolor{MediumVioletRed}{rgb}{0.78,0.08,0.52}
\definecolor{MidnightBlue}{rgb}{0.10,0.10,0.44}
\definecolor{MintCream}{rgb}{0.96,1.00,0.98}
\definecolor{MistyRose1}{rgb}{1.00,0.89,0.88}
\definecolor{MistyRose2}{rgb}{0.93,0.84,0.82}
\definecolor{MistyRose3}{rgb}{0.80,0.72,0.71}
\definecolor{MistyRose4}{rgb}{0.55,0.49,0.48}
\definecolor{MistyRose}{rgb}{1.00,0.89,0.88}
\definecolor{NavajoWhite1}{rgb}{1.00,0.87,0.68}
\definecolor{NavajoWhite2}{rgb}{0.93,0.81,0.63}
\definecolor{NavajoWhite3}{rgb}{0.80,0.70,0.55}
\definecolor{NavajoWhite4}{rgb}{0.55,0.47,0.37}
\definecolor{NavajoWhite}{rgb}{1.00,0.87,0.68}
\definecolor{NavyBlue}{rgb}{0.00,0.00,0.50}
\definecolor{OldLace}{rgb}{0.99,0.96,0.90}
\definecolor{OliveDrab1}{rgb}{0.75,1.00,0.24}
\definecolor{OliveDrab2}{rgb}{0.70,0.93,0.23}
\definecolor{OliveDrab3}{rgb}{0.60,0.80,0.20}
\definecolor{OliveDrab4}{rgb}{0.41,0.55,0.13}
\definecolor{OliveDrab}{rgb}{0.42,0.56,0.14}
\definecolor{OrangeRed1}{rgb}{1.00,0.27,0.00}
\definecolor{OrangeRed2}{rgb}{0.93,0.25,0.00}
\definecolor{OrangeRed3}{rgb}{0.80,0.22,0.00}
\definecolor{OrangeRed4}{rgb}{0.55,0.15,0.00}
\definecolor{OrangeRed}{rgb}{1.00,0.27,0.00}
\definecolor{PaleGoldenrod}{rgb}{0.93,0.91,0.67}
\definecolor{PaleGreen1}{rgb}{0.60,1.00,0.60}
\definecolor{PaleGreen2}{rgb}{0.56,0.93,0.56}
\definecolor{PaleGreen3}{rgb}{0.49,0.80,0.49}
\definecolor{PaleGreen4}{rgb}{0.33,0.55,0.33}
\definecolor{PaleGreen}{rgb}{0.60,0.98,0.60}
\definecolor{PaleTurquoise1}{rgb}{0.73,1.00,1.00}
\definecolor{PaleTurquoise2}{rgb}{0.68,0.93,0.93}
\definecolor{PaleTurquoise3}{rgb}{0.59,0.80,0.80}
\definecolor{PaleTurquoise4}{rgb}{0.40,0.55,0.55}
\definecolor{PaleTurquoise}{rgb}{0.69,0.93,0.93}
\definecolor{PaleVioletRed1}{rgb}{1.00,0.51,0.67}
\definecolor{PaleVioletRed2}{rgb}{0.93,0.47,0.62}
\definecolor{PaleVioletRed3}{rgb}{0.80,0.41,0.54}
\definecolor{PaleVioletRed4}{rgb}{0.55,0.28,0.36}
\definecolor{PaleVioletRed}{rgb}{0.86,0.44,0.58}
\definecolor{PapayaWhip}{rgb}{1.00,0.94,0.84}
\definecolor{PeachPuff1}{rgb}{1.00,0.85,0.73}
\definecolor{PeachPuff2}{rgb}{0.93,0.80,0.68}
\definecolor{PeachPuff3}{rgb}{0.80,0.69,0.58}
\definecolor{PeachPuff4}{rgb}{0.55,0.47,0.40}
\definecolor{PeachPuff}{rgb}{1.00,0.85,0.73}
\definecolor{PowderBlue}{rgb}{0.69,0.88,0.90}
\definecolor{RosyBrown1}{rgb}{1.00,0.76,0.76}
\definecolor{RosyBrown2}{rgb}{0.93,0.71,0.71}
\definecolor{RosyBrown3}{rgb}{0.80,0.61,0.61}
\definecolor{RosyBrown4}{rgb}{0.55,0.41,0.41}
\definecolor{RosyBrown}{rgb}{0.74,0.56,0.56}
\definecolor{RoyalBlue1}{rgb}{0.28,0.46,1.00}
\definecolor{RoyalBlue2}{rgb}{0.26,0.43,0.93}
\definecolor{RoyalBlue3}{rgb}{0.23,0.37,0.80}
\definecolor{RoyalBlue4}{rgb}{0.15,0.25,0.55}
\definecolor{RoyalBlue}{rgb}{0.25,0.41,0.88}
\definecolor{SaddleBrown}{rgb}{0.55,0.27,0.07}
\definecolor{SandyBrown}{rgb}{0.96,0.64,0.38}
\definecolor{SeaGreen1}{rgb}{0.33,1.00,0.62}
\definecolor{SeaGreen2}{rgb}{0.31,0.93,0.58}
\definecolor{SeaGreen3}{rgb}{0.26,0.80,0.50}
\definecolor{SeaGreen4}{rgb}{0.18,0.55,0.34}
\definecolor{SeaGreen}{rgb}{0.18,0.55,0.34}
\definecolor{SkyBlue1}{rgb}{0.53,0.81,1.00}
\definecolor{SkyBlue2}{rgb}{0.49,0.75,0.93}
\definecolor{SkyBlue3}{rgb}{0.42,0.65,0.80}
\definecolor{SkyBlue4}{rgb}{0.29,0.44,0.55}
\definecolor{SkyBlue}{rgb}{0.53,0.81,0.92}
\definecolor{SlateBlue1}{rgb}{0.51,0.44,1.00}
\definecolor{SlateBlue2}{rgb}{0.48,0.40,0.93}
\definecolor{SlateBlue3}{rgb}{0.41,0.35,0.80}
\definecolor{SlateBlue4}{rgb}{0.28,0.24,0.55}
\definecolor{SlateBlue}{rgb}{0.42,0.35,0.80}
\definecolor{SlateGray1}{rgb}{0.78,0.89,1.00}
\definecolor{SlateGray2}{rgb}{0.73,0.83,0.93}
\definecolor{SlateGray3}{rgb}{0.62,0.71,0.80}
\definecolor{SlateGray4}{rgb}{0.42,0.48,0.55}
\definecolor{SlateGray}{rgb}{0.44,0.50,0.56}
\definecolor{SlateGrey}{rgb}{0.44,0.50,0.56}
\definecolor{SpringGreen1}{rgb}{0.00,1.00,0.50}
\definecolor{SpringGreen2}{rgb}{0.00,0.93,0.46}
\definecolor{SpringGreen3}{rgb}{0.00,0.80,0.40}
\definecolor{SpringGreen4}{rgb}{0.00,0.55,0.27}
\definecolor{SpringGreen}{rgb}{0.00,1.00,0.50}
\definecolor{SteelBlue1}{rgb}{0.39,0.72,1.00}
\definecolor{SteelBlue2}{rgb}{0.36,0.67,0.93}
\definecolor{SteelBlue3}{rgb}{0.31,0.58,0.80}
\definecolor{SteelBlue4}{rgb}{0.21,0.39,0.55}
\definecolor{SteelBlue}{rgb}{0.27,0.51,0.71}
\definecolor{VioletRed1}{rgb}{1.00,0.24,0.59}
\definecolor{VioletRed2}{rgb}{0.93,0.23,0.55}
\definecolor{VioletRed3}{rgb}{0.80,0.20,0.47}
\definecolor{VioletRed4}{rgb}{0.55,0.13,0.32}
\definecolor{VioletRed}{rgb}{0.82,0.13,0.56}
\definecolor{WhiteSmoke}{rgb}{0.96,0.96,0.96}
\definecolor{YellowGreen}{rgb}{0.60,0.80,0.20}
\definecolor{aliceblue}{rgb}{0.94,0.97,1.00}
\definecolor{antiquewhite}{rgb}{0.98,0.92,0.84}
\definecolor{aquamarine1}{rgb}{0.50,1.00,0.83}
\definecolor{aquamarine2}{rgb}{0.46,0.93,0.78}
\definecolor{aquamarine3}{rgb}{0.40,0.80,0.67}
\definecolor{aquamarine4}{rgb}{0.27,0.55,0.45}
\definecolor{aquamarine}{rgb}{0.50,1.00,0.83}
\definecolor{azure1}{rgb}{0.94,1.00,1.00}
\definecolor{azure2}{rgb}{0.88,0.93,0.93}
\definecolor{azure3}{rgb}{0.76,0.80,0.80}
\definecolor{azure4}{rgb}{0.51,0.55,0.55}
\definecolor{azure}{rgb}{0.94,1.00,1.00}
\definecolor{beige}{rgb}{0.96,0.96,0.86}
\definecolor{bisque1}{rgb}{1.00,0.89,0.77}
\definecolor{bisque2}{rgb}{0.93,0.84,0.72}
\definecolor{bisque3}{rgb}{0.80,0.72,0.62}
\definecolor{bisque4}{rgb}{0.55,0.49,0.42}
\definecolor{bisque}{rgb}{1.00,0.89,0.77}
\definecolor{black}{rgb}{0.00,0.00,0.00}
\definecolor{blanchedalmond}{rgb}{1.00,0.92,0.80}
\definecolor{blue1}{rgb}{0.00,0.00,1.00}
\definecolor{blue2}{rgb}{0.00,0.00,0.93}
\definecolor{blue3}{rgb}{0.00,0.00,0.80}
\definecolor{blue4}{rgb}{0.00,0.00,0.55}
\definecolor{blueviolet}{rgb}{0.54,0.17,0.89}
\definecolor{blue}{rgb}{0.00,0.00,1.00}
\definecolor{brown1}{rgb}{1.00,0.25,0.25}
\definecolor{brown2}{rgb}{0.93,0.23,0.23}
\definecolor{brown3}{rgb}{0.80,0.20,0.20}
\definecolor{brown4}{rgb}{0.55,0.14,0.14}
\definecolor{brown}{rgb}{0.65,0.16,0.16}
\definecolor{burlywood1}{rgb}{1.00,0.83,0.61}
\definecolor{burlywood2}{rgb}{0.93,0.77,0.57}
\definecolor{burlywood3}{rgb}{0.80,0.67,0.49}
\definecolor{burlywood4}{rgb}{0.55,0.45,0.33}
\definecolor{burlywood}{rgb}{0.87,0.72,0.53}
\definecolor{cadetblue}{rgb}{0.37,0.62,0.63}
\definecolor{chartreuse1}{rgb}{0.50,1.00,0.00}
\definecolor{chartreuse2}{rgb}{0.46,0.93,0.00}
\definecolor{chartreuse3}{rgb}{0.40,0.80,0.00}
\definecolor{chartreuse4}{rgb}{0.27,0.55,0.00}
\definecolor{chartreuse}{rgb}{0.50,1.00,0.00}
\definecolor{chocolate1}{rgb}{1.00,0.50,0.14}
\definecolor{chocolate2}{rgb}{0.93,0.46,0.13}
\definecolor{chocolate3}{rgb}{0.80,0.40,0.11}
\definecolor{chocolate4}{rgb}{0.55,0.27,0.07}
\definecolor{chocolate}{rgb}{0.82,0.41,0.12}
\definecolor{coral1}{rgb}{1.00,0.45,0.34}
\definecolor{coral2}{rgb}{0.93,0.42,0.31}
\definecolor{coral3}{rgb}{0.80,0.36,0.27}
\definecolor{coral4}{rgb}{0.55,0.24,0.18}
\definecolor{coral}{rgb}{1.00,0.50,0.31}
\definecolor{cornflowerblue}{rgb}{0.39,0.58,0.93}
\definecolor{cornsilk1}{rgb}{1.00,0.97,0.86}
\definecolor{cornsilk2}{rgb}{0.93,0.91,0.80}
\definecolor{cornsilk3}{rgb}{0.80,0.78,0.69}
\definecolor{cornsilk4}{rgb}{0.55,0.53,0.47}
\definecolor{cornsilk}{rgb}{1.00,0.97,0.86}
\definecolor{cyan1}{rgb}{0.00,1.00,1.00}
\definecolor{cyan2}{rgb}{0.00,0.93,0.93}
\definecolor{cyan3}{rgb}{0.00,0.80,0.80}
\definecolor{cyan4}{rgb}{0.00,0.55,0.55}
\definecolor{cyan}{rgb}{0.00,1.00,1.00}
\definecolor{darkblue}{rgb}{0.00,0.00,0.55}
\definecolor{darkcyan}{rgb}{0.00,0.55,0.55}
\definecolor{darkgoldenrod}{rgb}{0.72,0.53,0.04}
\definecolor{darkgray}{rgb}{0.66,0.66,0.66}
\definecolor{darkgreen}{rgb}{0.00,0.39,0.00}
\definecolor{darkgrey}{rgb}{0.66,0.66,0.66}
\definecolor{darkkhaki}{rgb}{0.74,0.72,0.42}
\definecolor{darkmagenta}{rgb}{0.55,0.00,0.55}
\definecolor{darkolive}{rgb}{0.33,0.42,0.18}
\definecolor{darkorange}{rgb}{1.00,0.55,0.00}
\definecolor{darkorchid}{rgb}{0.60,0.20,0.80}
\definecolor{darkred}{rgb}{0.55,0.00,0.00}
\definecolor{darksalmon}{rgb}{0.91,0.59,0.48}
\definecolor{darksea}{rgb}{0.56,0.74,0.56}
\definecolor{darkslate}{rgb}{0.18,0.31,0.31}
\definecolor{darkslate}{rgb}{0.18,0.31,0.31}
\definecolor{darkslate}{rgb}{0.28,0.24,0.55}
\definecolor{darkturquoise}{rgb}{0.00,0.81,0.82}
\definecolor{darkviolet}{rgb}{0.58,0.00,0.83}
\definecolor{deeppink}{rgb}{1.00,0.08,0.58}
\definecolor{deepsky}{rgb}{0.00,0.75,1.00}
\definecolor{dimgray}{rgb}{0.41,0.41,0.41}
\definecolor{dimgrey}{rgb}{0.41,0.41,0.41}
\definecolor{dodgerblue}{rgb}{0.12,0.56,1.00}
\definecolor{firebrick1}{rgb}{1.00,0.19,0.19}
\definecolor{firebrick2}{rgb}{0.93,0.17,0.17}
\definecolor{firebrick3}{rgb}{0.80,0.15,0.15}
\definecolor{firebrick4}{rgb}{0.55,0.10,0.10}
\definecolor{firebrick}{rgb}{0.70,0.13,0.13}
\definecolor{floralwhite}{rgb}{1.00,0.98,0.94}
\definecolor{forestgreen}{rgb}{0.13,0.55,0.13}
\definecolor{gainsboro}{rgb}{0.86,0.86,0.86}
\definecolor{ghostwhite}{rgb}{0.97,0.97,1.00}
\definecolor{gold1}{rgb}{1.00,0.84,0.00}
\definecolor{gold2}{rgb}{0.93,0.79,0.00}
\definecolor{gold3}{rgb}{0.80,0.68,0.00}
\definecolor{gold4}{rgb}{0.55,0.46,0.00}
\definecolor{goldenrod1}{rgb}{1.00,0.76,0.15}
\definecolor{goldenrod2}{rgb}{0.93,0.71,0.13}
\definecolor{goldenrod3}{rgb}{0.80,0.61,0.11}
\definecolor{goldenrod4}{rgb}{0.55,0.41,0.08}
\definecolor{goldenrod}{rgb}{0.85,0.65,0.13}
\definecolor{gold}{rgb}{1.00,0.84,0.00}
\definecolor{gray0}{rgb}{0.00,0.00,0.00}
\definecolor{gray100}{rgb}{1.00,1.00,1.00}
\definecolor{gray10}{rgb}{0.10,0.10,0.10}
\definecolor{gray11}{rgb}{0.11,0.11,0.11}
\definecolor{gray12}{rgb}{0.12,0.12,0.12}
\definecolor{gray13}{rgb}{0.13,0.13,0.13}
\definecolor{gray14}{rgb}{0.14,0.14,0.14}
\definecolor{gray15}{rgb}{0.15,0.15,0.15}
\definecolor{gray16}{rgb}{0.16,0.16,0.16}
\definecolor{gray17}{rgb}{0.17,0.17,0.17}
\definecolor{gray18}{rgb}{0.18,0.18,0.18}
\definecolor{gray19}{rgb}{0.19,0.19,0.19}
\definecolor{gray1}{rgb}{0.01,0.01,0.01}
\definecolor{gray20}{rgb}{0.20,0.20,0.20}
\definecolor{gray21}{rgb}{0.21,0.21,0.21}
\definecolor{gray22}{rgb}{0.22,0.22,0.22}
\definecolor{gray23}{rgb}{0.23,0.23,0.23}
\definecolor{gray24}{rgb}{0.24,0.24,0.24}
\definecolor{gray25}{rgb}{0.25,0.25,0.25}
\definecolor{gray26}{rgb}{0.26,0.26,0.26}
\definecolor{gray27}{rgb}{0.27,0.27,0.27}
\definecolor{gray28}{rgb}{0.28,0.28,0.28}
\definecolor{gray29}{rgb}{0.29,0.29,0.29}
\definecolor{gray2}{rgb}{0.02,0.02,0.02}
\definecolor{gray30}{rgb}{0.30,0.30,0.30}
\definecolor{gray31}{rgb}{0.31,0.31,0.31}
\definecolor{gray32}{rgb}{0.32,0.32,0.32}
\definecolor{gray33}{rgb}{0.33,0.33,0.33}
\definecolor{gray34}{rgb}{0.34,0.34,0.34}
\definecolor{gray35}{rgb}{0.35,0.35,0.35}
\definecolor{gray36}{rgb}{0.36,0.36,0.36}
\definecolor{gray37}{rgb}{0.37,0.37,0.37}
\definecolor{gray38}{rgb}{0.38,0.38,0.38}
\definecolor{gray39}{rgb}{0.39,0.39,0.39}
\definecolor{gray3}{rgb}{0.03,0.03,0.03}
\definecolor{gray40}{rgb}{0.40,0.40,0.40}
\definecolor{gray41}{rgb}{0.41,0.41,0.41}
\definecolor{gray42}{rgb}{0.42,0.42,0.42}
\definecolor{gray43}{rgb}{0.43,0.43,0.43}
\definecolor{gray44}{rgb}{0.44,0.44,0.44}
\definecolor{gray45}{rgb}{0.45,0.45,0.45}
\definecolor{gray46}{rgb}{0.46,0.46,0.46}
\definecolor{gray47}{rgb}{0.47,0.47,0.47}
\definecolor{gray48}{rgb}{0.48,0.48,0.48}
\definecolor{gray49}{rgb}{0.49,0.49,0.49}
\definecolor{gray4}{rgb}{0.04,0.04,0.04}
\definecolor{gray50}{rgb}{0.50,0.50,0.50}
\definecolor{gray51}{rgb}{0.51,0.51,0.51}
\definecolor{gray52}{rgb}{0.52,0.52,0.52}
\definecolor{gray53}{rgb}{0.53,0.53,0.53}
\definecolor{gray54}{rgb}{0.54,0.54,0.54}
\definecolor{gray55}{rgb}{0.55,0.55,0.55}
\definecolor{gray56}{rgb}{0.56,0.56,0.56}
\definecolor{gray57}{rgb}{0.57,0.57,0.57}
\definecolor{gray58}{rgb}{0.58,0.58,0.58}
\definecolor{gray59}{rgb}{0.59,0.59,0.59}
\definecolor{gray5}{rgb}{0.05,0.05,0.05}
\definecolor{gray60}{rgb}{0.60,0.60,0.60}
\definecolor{gray61}{rgb}{0.61,0.61,0.61}
\definecolor{gray62}{rgb}{0.62,0.62,0.62}
\definecolor{gray63}{rgb}{0.63,0.63,0.63}
\definecolor{gray64}{rgb}{0.64,0.64,0.64}
\definecolor{gray65}{rgb}{0.65,0.65,0.65}
\definecolor{gray66}{rgb}{0.66,0.66,0.66}
\definecolor{gray67}{rgb}{0.67,0.67,0.67}
\definecolor{gray68}{rgb}{0.68,0.68,0.68}
\definecolor{gray69}{rgb}{0.69,0.69,0.69}
\definecolor{gray6}{rgb}{0.06,0.06,0.06}
\definecolor{gray70}{rgb}{0.70,0.70,0.70}
\definecolor{gray71}{rgb}{0.71,0.71,0.71}
\definecolor{gray72}{rgb}{0.72,0.72,0.72}
\definecolor{gray73}{rgb}{0.73,0.73,0.73}
\definecolor{gray74}{rgb}{0.74,0.74,0.74}
\definecolor{gray75}{rgb}{0.75,0.75,0.75}
\definecolor{gray76}{rgb}{0.76,0.76,0.76}
\definecolor{gray77}{rgb}{0.77,0.77,0.77}
\definecolor{gray78}{rgb}{0.78,0.78,0.78}
\definecolor{gray79}{rgb}{0.79,0.79,0.79}
\definecolor{gray7}{rgb}{0.07,0.07,0.07}
\definecolor{gray80}{rgb}{0.80,0.80,0.80}
\definecolor{gray81}{rgb}{0.81,0.81,0.81}
\definecolor{gray82}{rgb}{0.82,0.82,0.82}
\definecolor{gray83}{rgb}{0.83,0.83,0.83}
\definecolor{gray84}{rgb}{0.84,0.84,0.84}
\definecolor{gray85}{rgb}{0.85,0.85,0.85}
\definecolor{gray86}{rgb}{0.86,0.86,0.86}
\definecolor{gray87}{rgb}{0.87,0.87,0.87}
\definecolor{gray88}{rgb}{0.88,0.88,0.88}
\definecolor{gray89}{rgb}{0.89,0.89,0.89}
\definecolor{gray8}{rgb}{0.08,0.08,0.08}
\definecolor{gray90}{rgb}{0.90,0.90,0.90}
\definecolor{gray91}{rgb}{0.91,0.91,0.91}
\definecolor{gray92}{rgb}{0.92,0.92,0.92}
\definecolor{gray93}{rgb}{0.93,0.93,0.93}
\definecolor{gray94}{rgb}{0.94,0.94,0.94}
\definecolor{gray95}{rgb}{0.95,0.95,0.95}
\definecolor{gray96}{rgb}{0.96,0.96,0.96}
\definecolor{gray97}{rgb}{0.97,0.97,0.97}
\definecolor{gray98}{rgb}{0.98,0.98,0.98}
\definecolor{gray99}{rgb}{0.99,0.99,0.99}
\definecolor{gray9}{rgb}{0.09,0.09,0.09}
\definecolor{gray}{rgb}{0.75,0.75,0.75}
\definecolor{green1}{rgb}{0.00,1.00,0.00}
\definecolor{green2}{rgb}{0.00,0.93,0.00}
\definecolor{green3}{rgb}{0.00,0.80,0.00}
\definecolor{green4}{rgb}{0.00,0.55,0.00}
\definecolor{greenyellow}{rgb}{0.68,1.00,0.18}
\definecolor{green}{rgb}{0.00,1.00,0.00}
\definecolor{grey0}{rgb}{0.00,0.00,0.00}
\definecolor{grey100}{rgb}{1.00,1.00,1.00}
\definecolor{grey10}{rgb}{0.10,0.10,0.10}
\definecolor{grey11}{rgb}{0.11,0.11,0.11}
\definecolor{grey12}{rgb}{0.12,0.12,0.12}
\definecolor{grey13}{rgb}{0.13,0.13,0.13}
\definecolor{grey14}{rgb}{0.14,0.14,0.14}
\definecolor{grey15}{rgb}{0.15,0.15,0.15}
\definecolor{grey16}{rgb}{0.16,0.16,0.16}
\definecolor{grey17}{rgb}{0.17,0.17,0.17}
\definecolor{grey18}{rgb}{0.18,0.18,0.18}
\definecolor{grey19}{rgb}{0.19,0.19,0.19}
\definecolor{grey1}{rgb}{0.01,0.01,0.01}
\definecolor{grey20}{rgb}{0.20,0.20,0.20}
\definecolor{grey21}{rgb}{0.21,0.21,0.21}
\definecolor{grey22}{rgb}{0.22,0.22,0.22}
\definecolor{grey23}{rgb}{0.23,0.23,0.23}
\definecolor{grey24}{rgb}{0.24,0.24,0.24}
\definecolor{grey25}{rgb}{0.25,0.25,0.25}
\definecolor{grey26}{rgb}{0.26,0.26,0.26}
\definecolor{grey27}{rgb}{0.27,0.27,0.27}
\definecolor{grey28}{rgb}{0.28,0.28,0.28}
\definecolor{grey29}{rgb}{0.29,0.29,0.29}
\definecolor{grey2}{rgb}{0.02,0.02,0.02}
\definecolor{grey30}{rgb}{0.30,0.30,0.30}
\definecolor{grey31}{rgb}{0.31,0.31,0.31}
\definecolor{grey32}{rgb}{0.32,0.32,0.32}
\definecolor{grey33}{rgb}{0.33,0.33,0.33}
\definecolor{grey34}{rgb}{0.34,0.34,0.34}
\definecolor{grey35}{rgb}{0.35,0.35,0.35}
\definecolor{grey36}{rgb}{0.36,0.36,0.36}
\definecolor{grey37}{rgb}{0.37,0.37,0.37}
\definecolor{grey38}{rgb}{0.38,0.38,0.38}
\definecolor{grey39}{rgb}{0.39,0.39,0.39}
\definecolor{grey3}{rgb}{0.03,0.03,0.03}
\definecolor{grey40}{rgb}{0.40,0.40,0.40}
\definecolor{grey41}{rgb}{0.41,0.41,0.41}
\definecolor{grey42}{rgb}{0.42,0.42,0.42}
\definecolor{grey43}{rgb}{0.43,0.43,0.43}
\definecolor{grey44}{rgb}{0.44,0.44,0.44}
\definecolor{grey45}{rgb}{0.45,0.45,0.45}
\definecolor{grey46}{rgb}{0.46,0.46,0.46}
\definecolor{grey47}{rgb}{0.47,0.47,0.47}
\definecolor{grey48}{rgb}{0.48,0.48,0.48}
\definecolor{grey49}{rgb}{0.49,0.49,0.49}
\definecolor{grey4}{rgb}{0.04,0.04,0.04}
\definecolor{grey50}{rgb}{0.50,0.50,0.50}
\definecolor{grey51}{rgb}{0.51,0.51,0.51}
\definecolor{grey52}{rgb}{0.52,0.52,0.52}
\definecolor{grey53}{rgb}{0.53,0.53,0.53}
\definecolor{grey54}{rgb}{0.54,0.54,0.54}
\definecolor{grey55}{rgb}{0.55,0.55,0.55}
\definecolor{grey56}{rgb}{0.56,0.56,0.56}
\definecolor{grey57}{rgb}{0.57,0.57,0.57}
\definecolor{grey58}{rgb}{0.58,0.58,0.58}
\definecolor{grey59}{rgb}{0.59,0.59,0.59}
\definecolor{grey5}{rgb}{0.05,0.05,0.05}
\definecolor{grey60}{rgb}{0.60,0.60,0.60}
\definecolor{grey61}{rgb}{0.61,0.61,0.61}
\definecolor{grey62}{rgb}{0.62,0.62,0.62}
\definecolor{grey63}{rgb}{0.63,0.63,0.63}
\definecolor{grey64}{rgb}{0.64,0.64,0.64}
\definecolor{grey65}{rgb}{0.65,0.65,0.65}
\definecolor{grey66}{rgb}{0.66,0.66,0.66}
\definecolor{grey67}{rgb}{0.67,0.67,0.67}
\definecolor{grey68}{rgb}{0.68,0.68,0.68}
\definecolor{grey69}{rgb}{0.69,0.69,0.69}
\definecolor{grey6}{rgb}{0.06,0.06,0.06}
\definecolor{grey70}{rgb}{0.70,0.70,0.70}
\definecolor{grey71}{rgb}{0.71,0.71,0.71}
\definecolor{grey72}{rgb}{0.72,0.72,0.72}
\definecolor{grey73}{rgb}{0.73,0.73,0.73}
\definecolor{grey74}{rgb}{0.74,0.74,0.74}
\definecolor{grey75}{rgb}{0.75,0.75,0.75}
\definecolor{grey76}{rgb}{0.76,0.76,0.76}
\definecolor{grey77}{rgb}{0.77,0.77,0.77}
\definecolor{grey78}{rgb}{0.78,0.78,0.78}
\definecolor{grey79}{rgb}{0.79,0.79,0.79}
\definecolor{grey7}{rgb}{0.07,0.07,0.07}
\definecolor{grey80}{rgb}{0.80,0.80,0.80}
\definecolor{grey81}{rgb}{0.81,0.81,0.81}
\definecolor{grey82}{rgb}{0.82,0.82,0.82}
\definecolor{grey83}{rgb}{0.83,0.83,0.83}
\definecolor{grey84}{rgb}{0.84,0.84,0.84}
\definecolor{grey85}{rgb}{0.85,0.85,0.85}
\definecolor{grey86}{rgb}{0.86,0.86,0.86}
\definecolor{grey87}{rgb}{0.87,0.87,0.87}
\definecolor{grey88}{rgb}{0.88,0.88,0.88}
\definecolor{grey89}{rgb}{0.89,0.89,0.89}
\definecolor{grey8}{rgb}{0.08,0.08,0.08}
\definecolor{grey90}{rgb}{0.90,0.90,0.90}
\definecolor{grey91}{rgb}{0.91,0.91,0.91}
\definecolor{grey92}{rgb}{0.92,0.92,0.92}
\definecolor{grey93}{rgb}{0.93,0.93,0.93}
\definecolor{grey94}{rgb}{0.94,0.94,0.94}
\definecolor{grey95}{rgb}{0.95,0.95,0.95}
\definecolor{grey96}{rgb}{0.96,0.96,0.96}
\definecolor{grey97}{rgb}{0.97,0.97,0.97}
\definecolor{grey98}{rgb}{0.98,0.98,0.98}
\definecolor{grey99}{rgb}{0.99,0.99,0.99}
\definecolor{grey9}{rgb}{0.09,0.09,0.09}
\definecolor{grey}{rgb}{0.75,0.75,0.75}
\definecolor{honeydew1}{rgb}{0.94,1.00,0.94}
\definecolor{honeydew2}{rgb}{0.88,0.93,0.88}
\definecolor{honeydew3}{rgb}{0.76,0.80,0.76}
\definecolor{honeydew4}{rgb}{0.51,0.55,0.51}
\definecolor{honeydew}{rgb}{0.94,1.00,0.94}
\definecolor{hotpink}{rgb}{1.00,0.41,0.71}
\definecolor{indianred}{rgb}{0.80,0.36,0.36}
\definecolor{ivory1}{rgb}{1.00,1.00,0.94}
\definecolor{ivory2}{rgb}{0.93,0.93,0.88}
\definecolor{ivory3}{rgb}{0.80,0.80,0.76}
\definecolor{ivory4}{rgb}{0.55,0.55,0.51}
\definecolor{ivory}{rgb}{1.00,1.00,0.94}
\definecolor{khaki1}{rgb}{1.00,0.96,0.56}
\definecolor{khaki2}{rgb}{0.93,0.90,0.52}
\definecolor{khaki3}{rgb}{0.80,0.78,0.45}
\definecolor{khaki4}{rgb}{0.55,0.53,0.31}
\definecolor{khaki}{rgb}{0.94,0.90,0.55}
\definecolor{lavenderblush}{rgb}{1.00,0.94,0.96}
\definecolor{lavender}{rgb}{0.90,0.90,0.98}
\definecolor{lawngreen}{rgb}{0.49,0.99,0.00}
\definecolor{lemonchiffon}{rgb}{1.00,0.98,0.80}
\definecolor{lightblue}{rgb}{0.68,0.85,0.90}
\definecolor{lightcoral}{rgb}{0.94,0.50,0.50}
\definecolor{lightcyan}{rgb}{0.88,1.00,1.00}
\definecolor{lightgoldenrod}{rgb}{0.93,0.87,0.51}
\definecolor{lightgoldenrod}{rgb}{0.98,0.98,0.82}
\definecolor{lightgray}{rgb}{0.83,0.83,0.83}
\definecolor{lightgreen}{rgb}{0.56,0.93,0.56}
\definecolor{lightgrey}{rgb}{0.83,0.83,0.83}
\definecolor{lightpink}{rgb}{1.00,0.71,0.76}
\definecolor{lightsalmon}{rgb}{1.00,0.63,0.48}
\definecolor{lightsea}{rgb}{0.13,0.70,0.67}
\definecolor{lightsky}{rgb}{0.53,0.81,0.98}
\definecolor{lightslate}{rgb}{0.47,0.53,0.60}
\definecolor{lightslate}{rgb}{0.47,0.53,0.60}
\definecolor{lightslate}{rgb}{0.52,0.44,1.00}
\definecolor{lightsteel}{rgb}{0.69,0.77,0.87}
\definecolor{lightyellow}{rgb}{1.00,1.00,0.88}
\definecolor{limegreen}{rgb}{0.20,0.80,0.20}
\definecolor{linen}{rgb}{0.98,0.94,0.90}
\definecolor{magenta1}{rgb}{1.00,0.00,1.00}
\definecolor{magenta2}{rgb}{0.93,0.00,0.93}
\definecolor{magenta3}{rgb}{0.80,0.00,0.80}
\definecolor{magenta4}{rgb}{0.55,0.00,0.55}
\definecolor{magenta}{rgb}{1.00,0.00,1.00}
\definecolor{maroon1}{rgb}{1.00,0.20,0.70}
\definecolor{maroon2}{rgb}{0.93,0.19,0.65}
\definecolor{maroon3}{rgb}{0.80,0.16,0.56}
\definecolor{maroon4}{rgb}{0.55,0.11,0.38}
\definecolor{maroon}{rgb}{0.69,0.19,0.38}
\definecolor{mediumaquamarine}{rgb}{0.40,0.80,0.67}
\definecolor{mediumblue}{rgb}{0.00,0.00,0.80}
\definecolor{mediumorchid}{rgb}{0.73,0.33,0.83}
\definecolor{mediumpurple}{rgb}{0.58,0.44,0.86}
\definecolor{mediumsea}{rgb}{0.24,0.70,0.44}
\definecolor{mediumslate}{rgb}{0.48,0.41,0.93}
\definecolor{mediumspring}{rgb}{0.00,0.98,0.60}
\definecolor{mediumturquoise}{rgb}{0.28,0.82,0.80}
\definecolor{mediumviolet}{rgb}{0.78,0.08,0.52}
\definecolor{midnightblue}{rgb}{0.10,0.10,0.44}
\definecolor{mintcream}{rgb}{0.96,1.00,0.98}
\definecolor{mistyrose}{rgb}{1.00,0.89,0.88}
\definecolor{moccasin}{rgb}{1.00,0.89,0.71}
\definecolor{navajowhite}{rgb}{1.00,0.87,0.68}
\definecolor{navyblue}{rgb}{0.00,0.00,0.50}
\definecolor{navy}{rgb}{0.00,0.00,0.50}
\definecolor{oldlace}{rgb}{0.99,0.96,0.90}
\definecolor{olivedrab}{rgb}{0.42,0.56,0.14}
\definecolor{orange1}{rgb}{1.00,0.65,0.00}
\definecolor{orange2}{rgb}{0.93,0.60,0.00}
\definecolor{orange3}{rgb}{0.80,0.52,0.00}
\definecolor{orange4}{rgb}{0.55,0.35,0.00}
\definecolor{orangered}{rgb}{1.00,0.27,0.00}
\definecolor{orange}{rgb}{1.00,0.65,0.00}
\definecolor{orchid1}{rgb}{1.00,0.51,0.98}
\definecolor{orchid2}{rgb}{0.93,0.48,0.91}
\definecolor{orchid3}{rgb}{0.80,0.41,0.79}
\definecolor{orchid4}{rgb}{0.55,0.28,0.54}
\definecolor{orchid}{rgb}{0.85,0.44,0.84}
\definecolor{palegoldenrod}{rgb}{0.93,0.91,0.67}
\definecolor{palegreen}{rgb}{0.60,0.98,0.60}
\definecolor{paleturquoise}{rgb}{0.69,0.93,0.93}
\definecolor{paleviolet}{rgb}{0.86,0.44,0.58}
\definecolor{papayawhip}{rgb}{1.00,0.94,0.84}
\definecolor{peachpuff}{rgb}{1.00,0.85,0.73}
\definecolor{peru}{rgb}{0.80,0.52,0.25}
\definecolor{pink1}{rgb}{1.00,0.71,0.77}
\definecolor{pink2}{rgb}{0.93,0.66,0.72}
\definecolor{pink3}{rgb}{0.80,0.57,0.62}
\definecolor{pink4}{rgb}{0.55,0.39,0.42}
\definecolor{pink}{rgb}{1.00,0.75,0.80}
\definecolor{plum1}{rgb}{1.00,0.73,1.00}
\definecolor{plum2}{rgb}{0.93,0.68,0.93}
\definecolor{plum3}{rgb}{0.80,0.59,0.80}
\definecolor{plum4}{rgb}{0.55,0.40,0.55}
\definecolor{plum}{rgb}{0.87,0.63,0.87}
\definecolor{powderblue}{rgb}{0.69,0.88,0.90}
\definecolor{purple1}{rgb}{0.61,0.19,1.00}
\definecolor{purple2}{rgb}{0.57,0.17,0.93}
\definecolor{purple3}{rgb}{0.49,0.15,0.80}
\definecolor{purple4}{rgb}{0.33,0.10,0.55}
\definecolor{purple}{rgb}{0.63,0.13,0.94}
\definecolor{red1}{rgb}{1.00,0.00,0.00}
\definecolor{red2}{rgb}{0.93,0.00,0.00}
\definecolor{red3}{rgb}{0.80,0.00,0.00}
\definecolor{red4}{rgb}{0.55,0.00,0.00}
\definecolor{red}{rgb}{1.00,0.00,0.00}
\definecolor{rosybrown}{rgb}{0.74,0.56,0.56}
\definecolor{royalblue}{rgb}{0.25,0.41,0.88}
\definecolor{saddlebrown}{rgb}{0.55,0.27,0.07}
\definecolor{salmon1}{rgb}{1.00,0.55,0.41}
\definecolor{salmon2}{rgb}{0.93,0.51,0.38}
\definecolor{salmon3}{rgb}{0.80,0.44,0.33}
\definecolor{salmon4}{rgb}{0.55,0.30,0.22}
\definecolor{salmon}{rgb}{0.98,0.50,0.45}
\definecolor{sandybrown}{rgb}{0.96,0.64,0.38}
\definecolor{seagreen}{rgb}{0.18,0.55,0.34}
\definecolor{seashell1}{rgb}{1.00,0.96,0.93}
\definecolor{seashell2}{rgb}{0.93,0.90,0.87}
\definecolor{seashell3}{rgb}{0.80,0.77,0.75}
\definecolor{seashell4}{rgb}{0.55,0.53,0.51}
\definecolor{seashell}{rgb}{1.00,0.96,0.93}
\definecolor{sienna1}{rgb}{1.00,0.51,0.28}
\definecolor{sienna2}{rgb}{0.93,0.47,0.26}
\definecolor{sienna3}{rgb}{0.80,0.41,0.22}
\definecolor{sienna4}{rgb}{0.55,0.28,0.15}
\definecolor{sienna}{rgb}{0.63,0.32,0.18}
\definecolor{skyblue}{rgb}{0.53,0.81,0.92}
\definecolor{slateblue}{rgb}{0.42,0.35,0.80}
\definecolor{slategray}{rgb}{0.44,0.50,0.56}
\definecolor{slategrey}{rgb}{0.44,0.50,0.56}
\definecolor{snow1}{rgb}{1.00,0.98,0.98}
\definecolor{snow2}{rgb}{0.93,0.91,0.91}
\definecolor{snow3}{rgb}{0.80,0.79,0.79}
\definecolor{snow4}{rgb}{0.55,0.54,0.54}
\definecolor{snow}{rgb}{1.00,0.98,0.98}
\definecolor{springgreen}{rgb}{0.00,1.00,0.50}
\definecolor{steelblue}{rgb}{0.27,0.51,0.71}
\definecolor{tan1}{rgb}{1.00,0.65,0.31}
\definecolor{tan2}{rgb}{0.93,0.60,0.29}
\definecolor{tan3}{rgb}{0.80,0.52,0.25}
\definecolor{tan4}{rgb}{0.55,0.35,0.17}
\definecolor{tan}{rgb}{0.82,0.71,0.55}
\definecolor{thistle1}{rgb}{1.00,0.88,1.00}
\definecolor{thistle2}{rgb}{0.93,0.82,0.93}
\definecolor{thistle3}{rgb}{0.80,0.71,0.80}
\definecolor{thistle4}{rgb}{0.55,0.48,0.55}
\definecolor{thistle}{rgb}{0.85,0.75,0.85}
\definecolor{tomato1}{rgb}{1.00,0.39,0.28}
\definecolor{tomato2}{rgb}{0.93,0.36,0.26}
\definecolor{tomato3}{rgb}{0.80,0.31,0.22}
\definecolor{tomato4}{rgb}{0.55,0.21,0.15}
\definecolor{tomato}{rgb}{1.00,0.39,0.28}
\definecolor{turquoise1}{rgb}{0.00,0.96,1.00}
\definecolor{turquoise2}{rgb}{0.00,0.90,0.93}
\definecolor{turquoise3}{rgb}{0.00,0.77,0.80}
\definecolor{turquoise4}{rgb}{0.00,0.53,0.55}
\definecolor{turquoise}{rgb}{0.25,0.88,0.82}
\definecolor{violetred}{rgb}{0.82,0.13,0.56}
\definecolor{violet}{rgb}{0.93,0.51,0.93}
\definecolor{wheat1}{rgb}{1.00,0.91,0.73}
\definecolor{wheat2}{rgb}{0.93,0.85,0.68}
\definecolor{wheat3}{rgb}{0.80,0.73,0.59}
\definecolor{wheat4}{rgb}{0.55,0.49,0.40}
\definecolor{wheat}{rgb}{0.96,0.87,0.70}
\definecolor{whitesmoke}{rgb}{0.96,0.96,0.96}
\definecolor{white}{rgb}{1.00,1.00,1.00}
\definecolor{yellow1}{rgb}{1.00,1.00,0.00}
\definecolor{yellow2}{rgb}{0.93,0.93,0.00}
\definecolor{yellow3}{rgb}{0.80,0.80,0.00}
\definecolor{yellow4}{rgb}{0.55,0.55,0.00}
\definecolor{yellowgreen}{rgb}{0.60,0.80,0.20}
\definecolor{yellow}{rgb}{1.00,1.00,0.00}

\usepackage{comment}
\usepackage{amssymb}
\usepackage{amsfonts}
\usepackage{subcaption}
\usepackage{algorithmic}
\usepackage{algorithm}
\usepackage{pifont}
\usepackage{booktabs}
\usepackage{lipsum}
\usepackage{array}
\newcolumntype{P}[1]{>{\centering\arraybackslash}p{#1}}

\newcommand{\LP}[1]{\textcolor{red}{}}

\newcommand{\longonly}[1]{}

\usepackage{adjustbox}
\usepackage{booktabs}

\newcolumntype{R}[2]{%
    >{\adjustbox{angle=#1,lap=\width-(#2)}\bgroup}%
    l%
    <{\egroup}%
}

\graphicspath{{figures/},{figures/new_diagrams/}}

\usepackage{hyperref}

\renewcommand{\baselinestretch}{0.97}
\usepackage{mdwlist}

\let\itemize\undefined
\makecompactlist{itemize}{stditemize}

\let\enumerate\undefined
\makecompactlist{enumerate}{stdenumerate}

\usepackage[tracking=false,kerning=true,spacing=true]{microtype}
\title{\LARGE \bf
Team NCTU: Toward AI-Driving for Autonomous Surface Vehicles - From Duckietown to RobotX}

\author{Yi-Wei Huang$^{1}$, Tzu-Kuan Chuang$^{1}$, Ni-Ching Lin$^{1}$ 
Yu-Chieh Hsiao$^{1}$, Pin-Wei Chen$^{1}$, Ching-Tang Hung$^{2}$,   \\ 
Shih-Hsing Liu$^{1}$, Hsiao-Sheng Chen$^{1}$, Ya-Hsiu Hsieh$^{1}$,  \\
Ching-Tang Hung$^{2}$, Yen-Hsiang Huang$^{2}$,    \\
Yu-Xuan Chen$^{3}$, Kuan-Lin Chen$^{3}$, Ya-Jou Lan$^{3}$,  \\
Chao-Chun Hsu$^{1}$, Chun-Yi Lin$^{1}$, Jhih-Ying Li$^{1}$, \\
Jui-Te Huang$^{1}$, Yu-Jen Menn$^{1}$, Sin-Kiat Lim$^{1}$, \\
Kim-Boon Lua$^{3}$, Chia-Hung Dylan Tsai$^{3}$, Chi-Fang Chen$^{2}$, and Hsueh-Cheng Wang$^{*, 1}$
\thanks{$^{1}$Department of Electrical and Computer Engineering, National Chiao
Tung University, Taiwan. Corresponding author email:
        {\tt\small hchengwang@g2.nctu.edu.tw}}%
\thanks{$^{2}$Department of Engineering Science and Ocean Engineering, National Taiwan University, Taiwan. }%
\thanks{$^{3}$Department of Mechanical Engineering, National Chiao
Tung University, Taiwan.}%
}

\begin{document}

\maketitle
\thispagestyle{empty}
\pagestyle{empty}


\begin{abstract}

Robotic software and hardware systems of autonomous surface vehicles have been developed in transportation, military, and ocean researches for decades. Previous efforts in RobotX Challenges 2014 and 2016 facilitates the developments for important tasks such as obstacle avoidance and docking. Team NCTU is motivated by the AI Driving Olympics (AI-DO) developed by the Duckietown community, and adopts the principles to RobotX challenge. With the containerization (Docker) and uniformed AI agent (with observations and actions), we could better 1) integrate solutions developed in different middlewares (ROS and MOOS), 2) develop essential functionalities of from simulation (Gazebo) to real robots (either miniaturized or full-sized WAM-V), and 3) compare different approaches either from classic model-based or learning-based. Finally, we setup an outdoor on-surface platform with localization services for evaluation. Some of the preliminary results will be presented for the Team NCTU participations of the RobotX competition in Hawaii in 2018.

\end{abstract}

\section{Introduction}

RobotX competition has brought together the top schools among the nations around oceans, and enhanced the developments of both hardware designs and software algorithms for perceptions and propulsions of unmanned surface vehicles. Perceptions play an important role together with state estimation and motion planning for 
autonomous vehicles. The participated teams have been working toward robustness to degradation caused by motion, scale and perspective transformation from different viewing positions, warp and occlusion, and variants of color from light condition, and speed and accuracy to support real-time decision-making. 

Nevertheless, Building an autonomous unmanned marine system is challenging in many aspects. The hardware should be robust enough to maintain functionalities in different weather conditions. Waterproof design, compact electronics, together with cooling system are crucial for weather conditions of heavy rain, strong winds and the blistering sun. Due to the resource constrained of power and computation, realtime algorithms should be adapted for overall performance. Those challenges remain due to the uncertainties of the marine environments, and the lack of baselines and standardized evaluations because benchmarking in marine environments is hard. More importantly, there are still big questions arisen from the previous RobotX competitions.

\begin{figure}[t]
  \centering
    \includegraphics[width=\columnwidth]{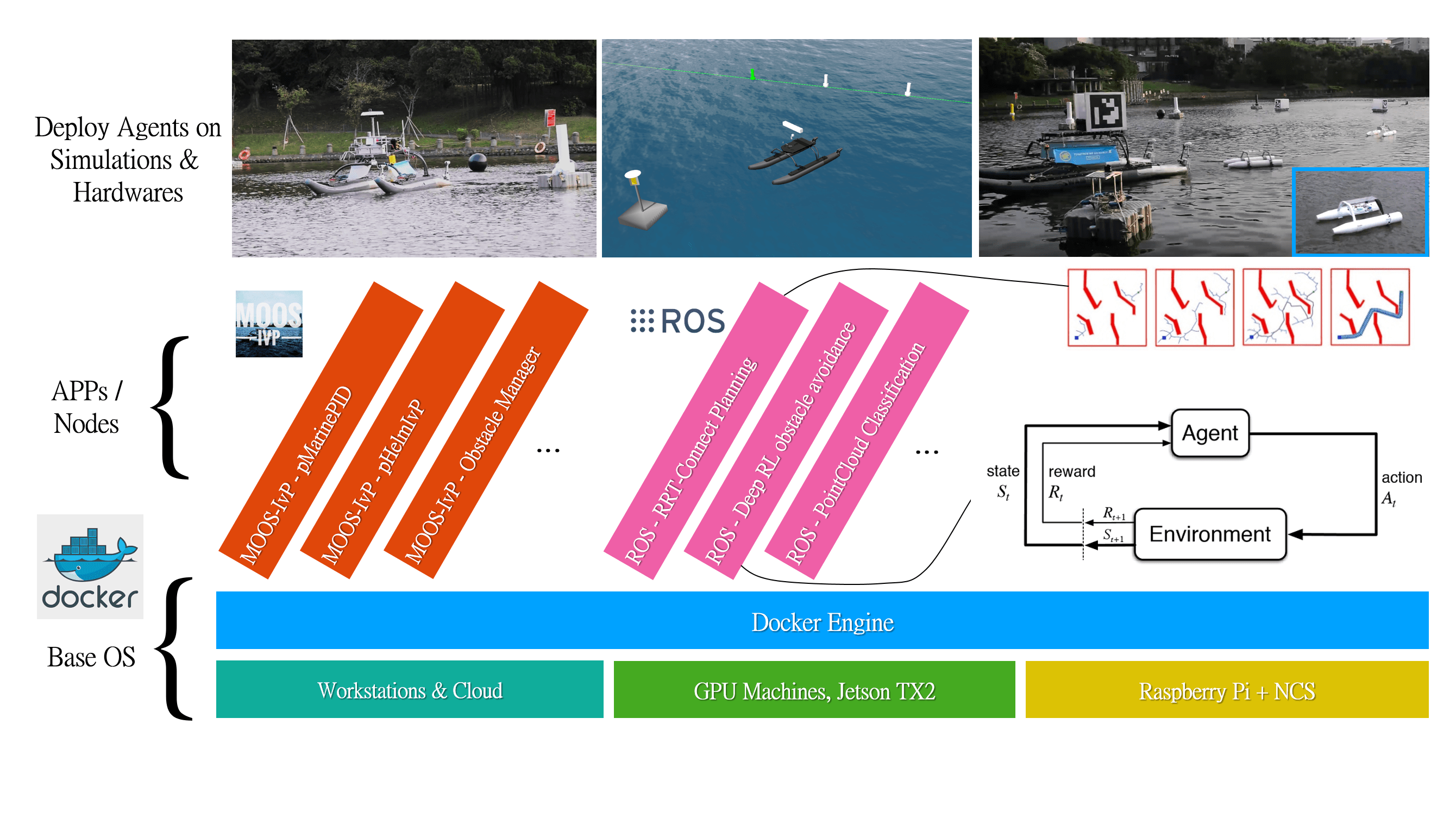}
        \caption{We proposed an autonomous surface/underwater system with modular design and is compatible both with ROS and MOOS. The system is capable of running in real robots, minitured robots and simulations. The capability of running in multiple platforms provide us better tools to develop deep learning algorithms.}
 \label{figure:teaser}
\end{figure}

\begin{enumerate}

\item What are the basic principles to manage multiple tasks and to advance the developments, instead of just a set of clever hacks on individual tasks?
\item Many of the tasks have some hand-tuned parameters that were optimized for the tasks during the competition. How could the developed system/techniques be generalized to other scenarios, on other robots, or in other environments?
\item Can we use simulation environments to facilitate the developments? What are the gaps between real and virtual environments and how to bring the gaps closer?  

\end{enumerate}

Team NCTU's technical approaches are motivated from the challenges and big questions above. The recent mega trends of AI have fostered researchers in robotics, machine learning, and other fields together. In particular, we wish to adopt the principles of the AI Driving Olympics (AI-DO)~\cite{aido, aido-duckietown} hosted in NIPS 2018 into RobotX competition. The AI-DO is developed by the Duckietown community. The project was initiated in MIT in 2016 and is now offered as university courses in ETH Zurich, University of Montreal, and Toyota Technological Institute at Chicago, and National Chiao Tung University (NCTU) in 2017. Duckietown ~\cite{duckietown_mit} is an open, reproducible, and inexpensive robotic education and research platform. A team of vehicles are built upon Robot Operation System (ROS) and include an onboard monocular camera and an embedded computer. A miniaturized city (Duckietown) with roads, signage, and obstacles is designed to tackle the problems of autonomy. The Duckietown platform started to embrace Docker and deep learning in 2018 and host the competitions of a few tasks via learning approaches such as 
Convolution Neural Network (CNN) and Deep Reinforcement Learning. Here we wish to transform such methodologies into the domain of unmanned surface vehicles. 

We summarize our contributions as follows:

\begin{enumerate}
\item
With the principles of containerization (Docker), we designed and built an autonomous unmanned marine system with the hardware and software that is compatible for two commonly-used middlewares: ROS and MOOS. The containerization allows to a) develop under different middlewares, b) deploy on real and simulation environments, and c)  modularize and compare algorithms for the RobotX tasks. 
\item 
The uniformed AI agent framework with observations and actions facilitate comparisons between classic and learning-based algorithms. In particular, we tackle the problems of a)   the placard detection and 3D object recognition carried out in classic feature-based vs. learning-based methods, b) obstacle avoidance via classical methods vs. deep reinforcement learning approaches.  
\item 
We built an outdoor on-surface motion capture system based on 3D LiDAR and 2D vision-based approaches as benchmarks for further research purposes.
\end{enumerate}

\begin{figure}[H]
  \centering
    \includegraphics[width=\columnwidth]{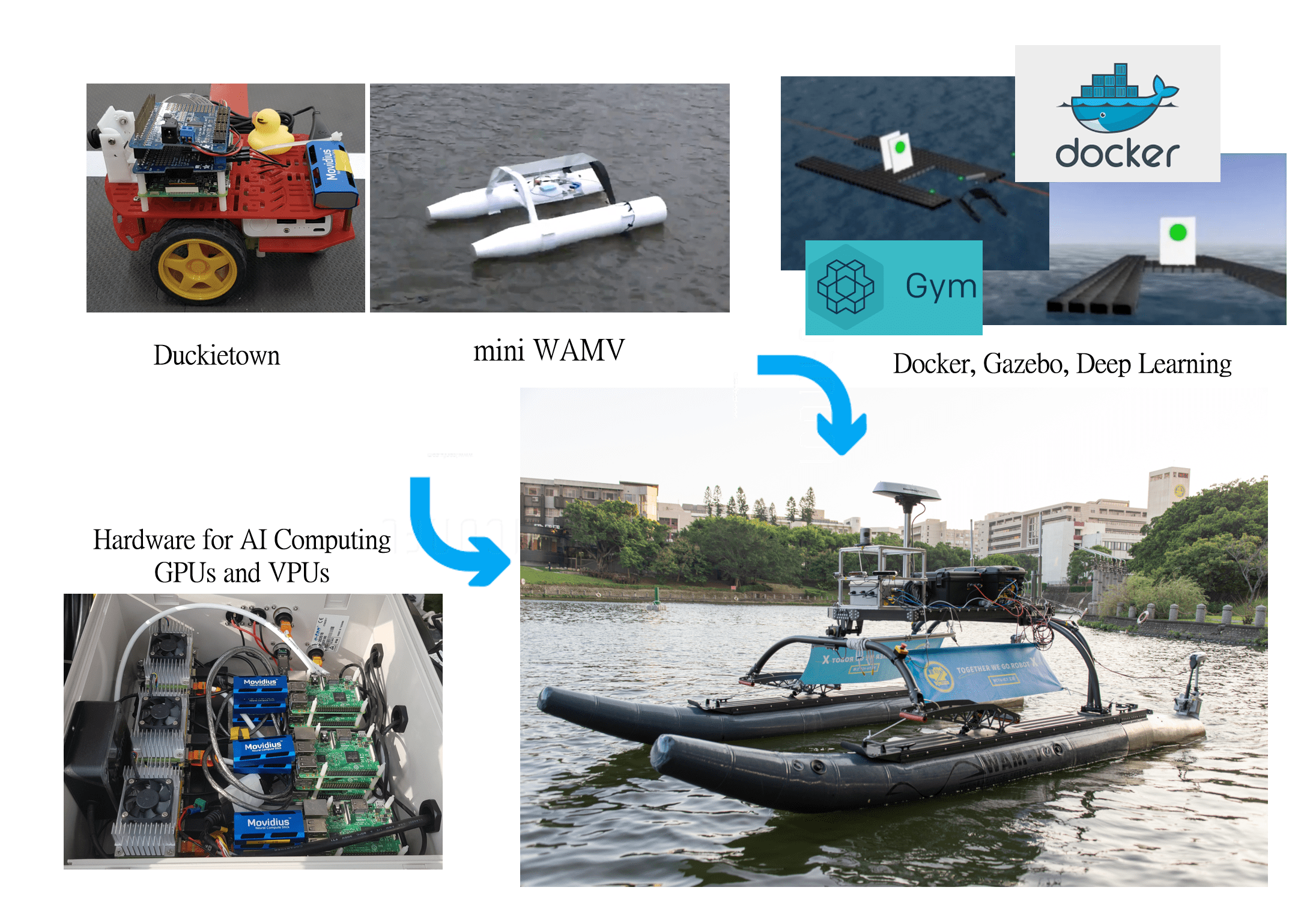}
        \caption{From duckietown to RobotX}
 \label{figure:motivation}
\end{figure}


\section{Literature Review}

\subsection{RobotX 2014 and 2016}
RobotX has already been held for 2 times in 2014 and 2016. The 2014 championship team MIT-Olin's~\cite{anderson2016overview} uses IvP-helm~\cite{benjamin2003interval} to achieve multi-objective optimization, such as transit to target waypoint while avoiding obstacles. The MOOS~\cite{benjamin2009overview} framework operates the speed and heading which are generated by IvP Helm and also process data, for instance, GPS, object map, PID controller, and so on. The 2016 championship team  Team NaviGator AMS~\cite{frank2016university} utilize ROS~\cite{quigley2009ros} as middleware. That year added a new mission about underwater shape identification. The team designed underwater vehicle Anglerfish~\cite{gray2016anglerfish} by themselves which can be controlled by the NaviGator ASV via a 30-metered tether providing power and ethernet. Most of the teams in the past two competitions apply the algorithms about detection and motion without using machine learning methods. 

\subsection{Learning Approaches for Mobile Robots}

Recently, deep Convolutional Neural Networks (CNN) have been used to achieve autonomous trail or lane following. Giusti et al. \cite{giusti2016machine} tackled autonomous forest or mountain trail-following using a single monocular camera mounted on a mobile robot, such as a micro-aerial vehicle. Unlike the previous literature, they focused on trail segmentation and used low-level features to develop a supervised learning approach using a deep CNN classifier. The trained CNN classifier was shown to follow unseen trails using a quadrotor. Deep driving~\cite{chen2015deepdriving} categorizes the autonomous driving work into three paradigms. \emph{Behavior Reflex} is known as a low-level approach for constructing a direct mapping from the image/sensory inputs to produce a steering motion. This is done by means of a deep CNN trained by labels generated from human driving along a road or in virtual environments. \emph{Mediated perception} is the recognition of driving-relevant objects, e.g., lanes, traffic signs, traffic lights, cars, or pedestrians. The recognition results are then combined into a consistent world representation of the cars and immediate surroundings. \emph{Direct perception} falls between mediated perception and behavior reflex. It proposes to learn a mapping from an image to estimate several meaningful states of the road situation, such as the angle of the car relative to the road and the lateral distance to lane markings. With the state estimation, other filters or FSMs and controllers can be applied. The forest-trail-following vehicle in \cite{giusti2016machine} belongs to behavior reflex, whereas the Duckietown falls into the direct perception paradigm.

In the context of deep learning, the most commonly used method for simulation to real environment is transfer learning~\cite{peng2015learning, su2015render}. By gathering data from the target domain in addition to the source domain closes the gap between simulation and the reality. \cite{tzeng2015towards, tzeng2015simultaneous} uses data alignment to tackle with simulation to real problems for robotics arm pose estimation. \cite{zhang2015towards, zhu2017target} proved that doing simple but precise tasks in simulations can be transferred to real world robots for more complex and general tasks. \cite{rusu2016progressive} connected the layers of deep learning models trained with simulation data and real world data together. It resulted features trained in virtual environments usable in real world scenarios.

\section{Design Strategy}

\subsection{Enabling Hardware Systems for AI Computing }
For the first time building such a sophisticated system, it is very crucial to do every possible evaluation beforehand. It just happens that we've been working on a project called "Duckietown", a miniaturized AI self-driving car platform that runs on ROS. The cars a.k.a "duckiebots" use AI to navigate around and communicate with other duckiebots. We think it would be interesting to apply the AI technology to our Autonomous Maritime System(AMS). The duckiebots could provide a miniaturized environment for testing our algorithms. 

\subsection{From Simulation to Real Environments}
In addition to the duckietown platform as an simulation, using software simulation is also important for our proof of concept. We use ROS as middleware, so Gazebo~\cite{koenig2004design} is the obvious solution due to the compatibility to ROS. Using the advantage of ROS, we could simply interface with simultation and real robots by publishing and subsribing corresponding messages. A few parameters should be adjusted to fit the real environment. 

\subsection{Containerized Algorithms for Deployable Softwares}
To deploy a variety of different algorithms on various environments, software dependencies may be troubling. A decent solution for it is to containerize algorithms. By using Docker, every algorithm is like a building block, which is easy to switch to one another. The plug and play feature on docker containers provide us with simple deployment on both simulation and real environments.

\subsection{Comparing Deep Learning to Classic Approaches}
Deep learning has influenced the robotic research in the past few years. The challenge has always been proving deep learning is better than classic approaches. We believe that both are good but in different aspects. Classic approaches may deal with a problem with a really great performance, on the other hand deep learning methods might be more reliable to unexpected environment changes. Our goal is to compare the two and discuss the pros and cons of each method.

\section{Vehicle Design}
	
\begin{figure}[t]
	  \centering
		\includegraphics[width=\columnwidth]{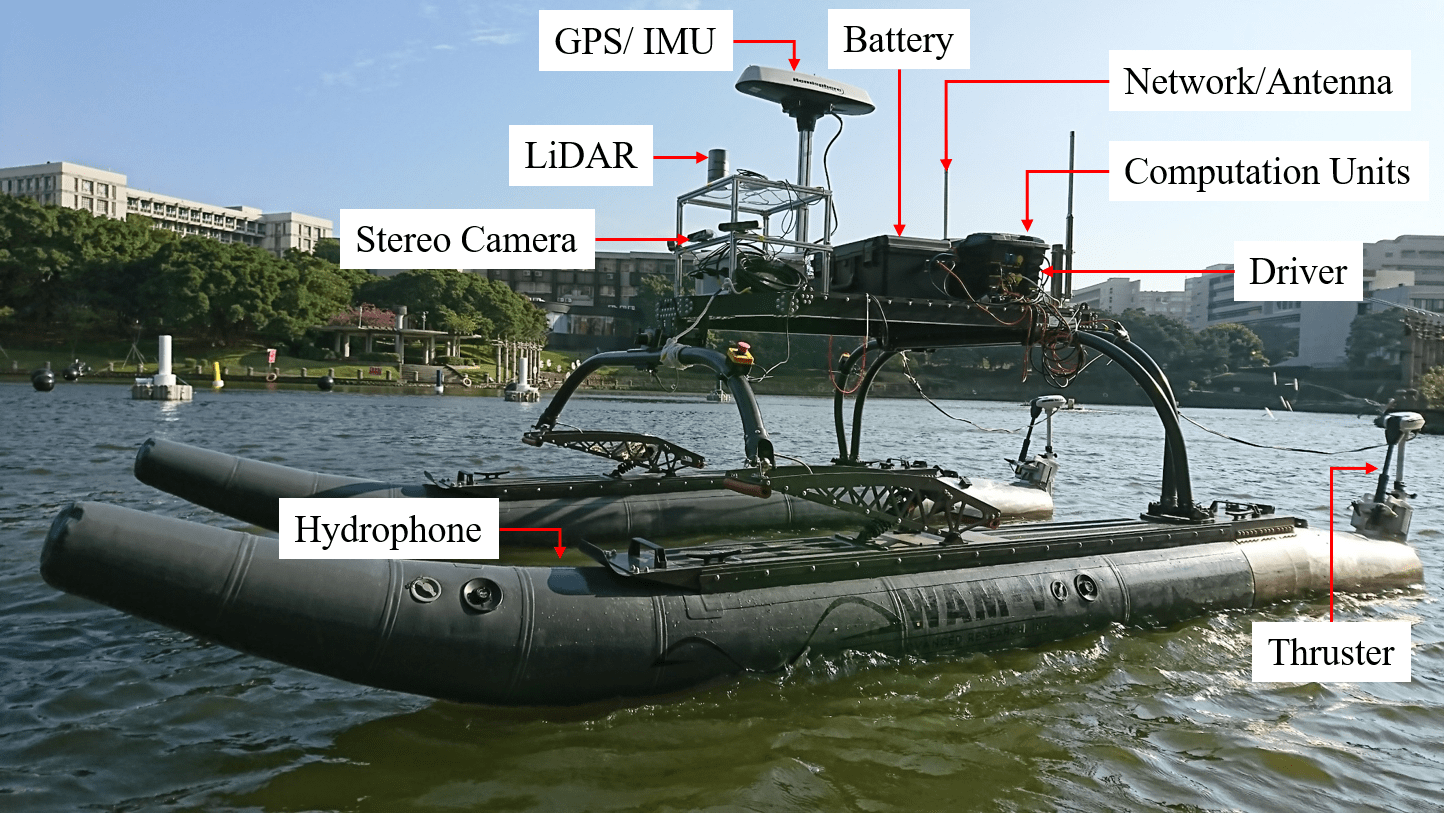}
			\caption{Hardware}
	 \label{figure:hardware_setting}
	\end{figure}

	The WAM-V system has one multi-enclosures on payload which can be seen in Fig.~\ref{figure:hardware_setting}, which split into two power enclosures, one sensor enclosure, one controller enclosure, and one hydronphone enclosure. 
	
	\subsection{Propulsion}
	
	Two forward thrusts are used for the differential drive system. Each motor is capable of 80 lbs of thrust. It is the simplest to implement with the least amount of cost. We once considered holonomic drive which enables the mobility of the USV, but it requires at least three thrusters and it has much more complicated control motion. In addition, we think there must be some reason that no holonomic drive USV is currently working in the ocean.
	
	\subsection{Sensor System}
	
	The sensor system is build as an sensor tower and an independent sensor for underwater acoustic array for the competition. The sensor tower has 4 levels, each mounted with different sensors. (shown as Fig.~\ref{figure:hardware_setting}) One 3D LiDAR, three depth cameras, an IMU, and a GPS are used to sense the environment. These sensors could localize our USV and target objects. There is also an underwater hydrophone array for detecting underwater pingers.

	\subsection{Computation}
	
	This paper presents a multi-computing-unit system. A Industrial Personal Computer(IPC) served as the main computing unit. Through ethernet it communicates with all other computing units, such as Nvidia Jetson TX2, and Raspberry Pi 3 for different purposes. The Nvidia Jetson TX2 is used to gather depth camera sensor data and sent back to the IPC. The Raspberry Pi 3 is used to control motors. Every unit in the system is connected by ethernet and communicate with ROS.
	
	\begin{table}[h]
	\vspace{0.2cm}
	\caption{Computation Units}
		\centering
		\begin{tabular} {|l|c|c|l|}
			\hline
			Computation Units			& \rotatebox{60}{Number}	& \rotatebox{60}{Processor}	& \rotatebox{60}{Spec/Used} \\
			\hline
			ASUS Laptop (GX501) 		& 1 		& GPU 		& GeForce GTX 1050 Ti \\
			\hline
			NVidia Jeson TX2 			& 3 		& GPU 		& Pascal 256 CUDA cores \\
			\hline
			Raspberry Pi 3 (Ne-			& 9			& VPU 		& Myriad 2 Vision\\
			ural Compute Stick)			& 			&  		& Processing Unit\\
			\hline
			Raspberry Pi 3 				& 1 		&  		& Propulsion \\
			\hline
			Raspberry Pi 3 				& 1 		&  		& Visual Feedback \\
			\hline
			Raspberry Pi 3 				& 1 		&  		& Launching and Recovery \\
			\hline
			Raspberry Pi 3 				& 1 		&  		& Detect and Deliver \\
			\hline
		\end{tabular}
		\label{table:computation_units}
	\end{table}
	







\section{Software system design}

\begin{figure*}[ht]
  \centering
    \includegraphics[width=0.8\textwidth]{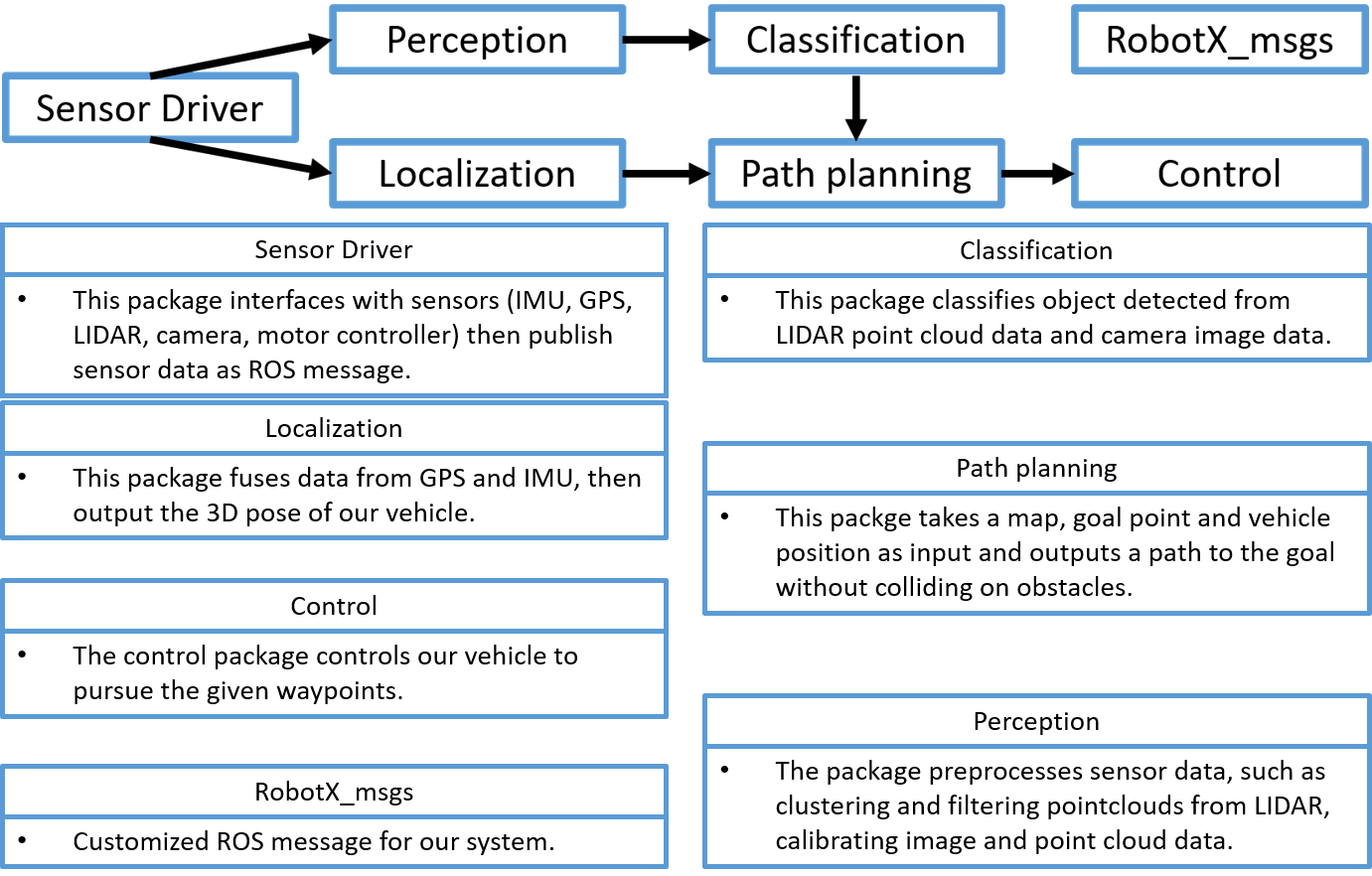}
        \caption{The software architecture}
 \label{figure:repo_arch}
\end{figure*}

\subsection{Overall System}
The software is built upon the ROS interface, several nodes are built for distinct functionalites and they communicate with each other via ROS messages and services. The ROS package tree contains nodes shown in Figure~\ref{figure:repo_arch}

There are nodes that deal with path planning, localization, control, perception, classification, etc. Some of the ROS nodes are contained in docker for easier deployment. 

\subsection{AI Agent}
To generalize the usage of this system, we introduce the concept of AI agent. The idea is to build it so its compatible for OpenAI gym~\cite{brockman2016openai}.
\begin{figure}[H]
  \centering
    \includegraphics[width=0.8\columnwidth]{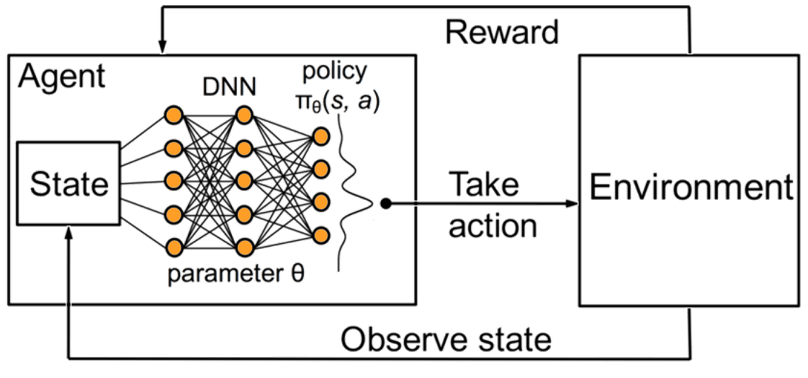}
        \caption{The relationship between the AI agent and the environment}
 \label{figure:ai_agent}
\end{figure}
Figure~\ref{figure:ai_agent} shows that the agent should output observations and rewards for each input action. The design of the agent should output higher rewards for correct actions. Then the algorithm tries to interact with the agent and learn what the AI agent wants it to learn. So for each task there is at least one AI agent and each agent could have different observation from various of sensors.

\subsection{MOOS and ROS}
The Mission Oriented Operating System (MOOS)~\cite{benjamin2009overview} and the Robot Operating System (ROS)~\cite{quigley2009ros} are both robotic frameworks which provide the communication between different robots and components. \par
ROS is an open source project which is compatible of using C++ and Python as development languages. In ROS, processing units are called "Nodes" and they communicate with each other by "Messages" and "Services". A central monitor program named "ROS Master" controls all the status of nodes and handles all the "messages" and "services" exchanged among the ROS "nodes". Since there are majority of users and thousands of modules have been developed, ROS already became the most commonly-used middleware for robotics. Lots of robots such as PR2, Atlas, UR5, Turtlebot use this open source software as their software framework. \par

MOOS-IvP is the combination of two open source projects: MOOS and Ivp Helm. MOOS is developed by the University of Oxford which and is designed to be the core of autonomy middleware. While IvP Helm is developed by MIT used for multi-objective optimization between competing behaviors. Most of the MOOS-IvP projects are applied in the field of unmanned surface vehicle and underwater acoustics and large influence in the field of marine robots for years. \par	
This paper~\cite{west2011overview} from Georgia Tech presented some advantages and disadvantages of these two middlewares. MOOS is more oriented towards onboard publish-subscribe architecture by using its community database "MOOSDB" and more lightweight. In addition, there are hundreds of executable behaviors, simulations, and MOOS Apps developed by the marine robotics community. This leads to easy development of your own algorithms for USVs. On the other hand, ROS is more general and provides multi-platform services which is less painful for system integration. Moreover, it has many of the low-level device control interfaces and components int the hardware abstraction layer. ROS also included Gazebo, a 3D simulation for general robotics application, on the contrary MOOS only offers 2D simulation. However, development in Gazebo is much complicated than in MOOS. \par	
We chose to use both frameworks for taking advantages from each. In order to cooperate the two frameworks, we use MOOS-ROS Bridge~\cite{demarco2011implementation} to communicate these two robotics middlewares. We modified and added a few features for better performance in applications needed by our system.

\subsection{Localization}
One of the critical things to do is to obtain the vessel's pose. Without the poses, the vessel wouldn't know where it is not to mention navigate to a position. We implement the localization with a Hector GPS and a microstrain 9 DoF IMU. GPS datum and IMU datum are fused together using a Gaussian filter. This filter stablizes the localizaton output. Posistion and Orientation are calculated separately.

\subsection{Control and Navigation}
Another crucial feature is to control the vessel and to navigate from one point to another. We designed our WAM-V's with a differential drive motion model, therefore linear and angular could be controlled separately. We use a PID controller for heading control and a cascade PID controller for the position. We chose the cascade PID controller to control position because with both position and velocity feedback, it could provide better performance for tasks i.e. station keeping. 

\begin{figure}[H]
  \centering
    \includegraphics[width=0.8\columnwidth]{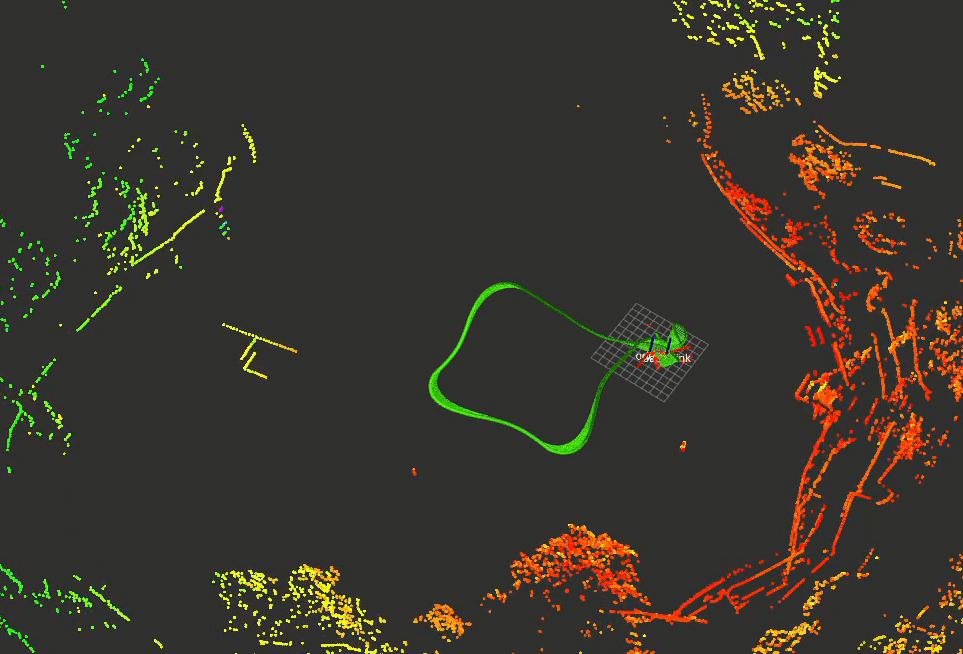}
        \caption{The green line shows trajectory of the vessel executing a square movement. Other colored points are the pointcloud from the LiDAR data.}
 \label{figure:square_nav}
\end{figure}

For navigation, we implement the pure pursuit algorithm for waypoint navigation. The pure pursuit algorith makes the trajectory smoother by reducing sharp turns. Figure~\ref{figure:square_nav} shows the trajectory for our WAM-V executing waypoint navigation. The waypoints are a the points of a square with a 15 meter edge. It show the turns on the corners are smoothed out.

\subsection{Object Classification}
\begin{figure}[h]
  \centering
    \includegraphics[width=0.6\columnwidth]{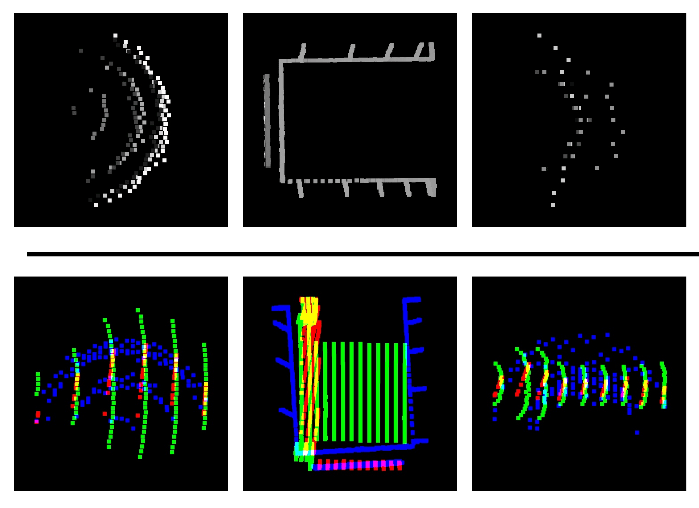}
        \caption{Each target object's pointcloud is projected to the X-Y, Y-Z and X-Z planes then the three flattened images saved to the R, G, B channels of a 2D-image}
 \label{figure:pcl_classification}
\end{figure}

Due to the precarious weather and changeable lighting conditions which can strongly impact the color of the object, we chose to use the pointclouds gathered from the LIDAR for object classification. A brief introduction of the algorithm pipeline is as follows. First, for the preprocessing stage, we apply RANSAC and noise filter to remove the points from the sea level and random noises, leaving the objects remaining in the pointcloud. After the preprocessing process, we apply a clustering algorithm to separate each object for each another. Then we project each object's pointcloud to X-Y, Y-Z and X-Z planes (according to the LIDAR's coordinate frame) and saved each plane projection to one of the RGB channels of an image. (Figure~\ref{figure:pcl_classification}) We now obtain the RGB image and named it as the "flattened pointcloud". This "flattened pointcloud" remains some of the 3D information and at the same time is a more compact datatpe compared to the raw pointcloud.

\begin{figure}[H]
  \centering
    \includegraphics[width=\columnwidth]{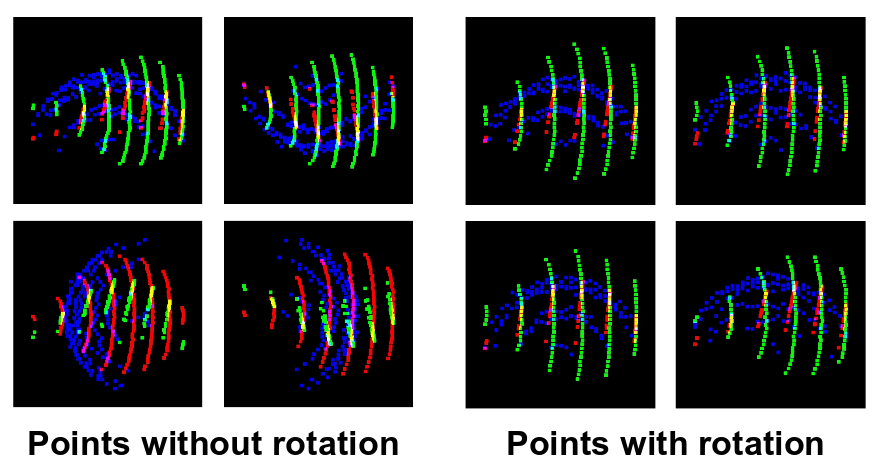}
        \caption{With the rotation technique, the flattened pointcloud from the same object but 4 different viewing angles looks almost identical compared to those with out the rotation process. This improves our classification performance.}
 \label{figure:pcl_classification_rot}
\end{figure}

However, the projection is related to the LiDAR's orientation, different rotations of the WAM-V cause the "flattened pointcloud" of the same object to differ a lot. This may lead to bad performance for classification afterwards. To cope with this problem, we simply transform the coordinate frame to the object's frame which sets the origin to the center of the object and the x axes. This resulted the flatted image be consistent as the WAM-V change the orientation. Figure~\ref{figure:pcl_classification_rot}.

For each class we gather approximately 700 "flatten pointcloud" images, with 500 images from Gazebo virtual environment and 200 images from the real world (Bamboo lake in NCTU, Taiwan). We have 5 different classes, which are obstacle buoys, totem buoys, the dock and the box for the detect and deliver task. 

\begin{figure}[H]
  \centering
    \includegraphics[width=\columnwidth]{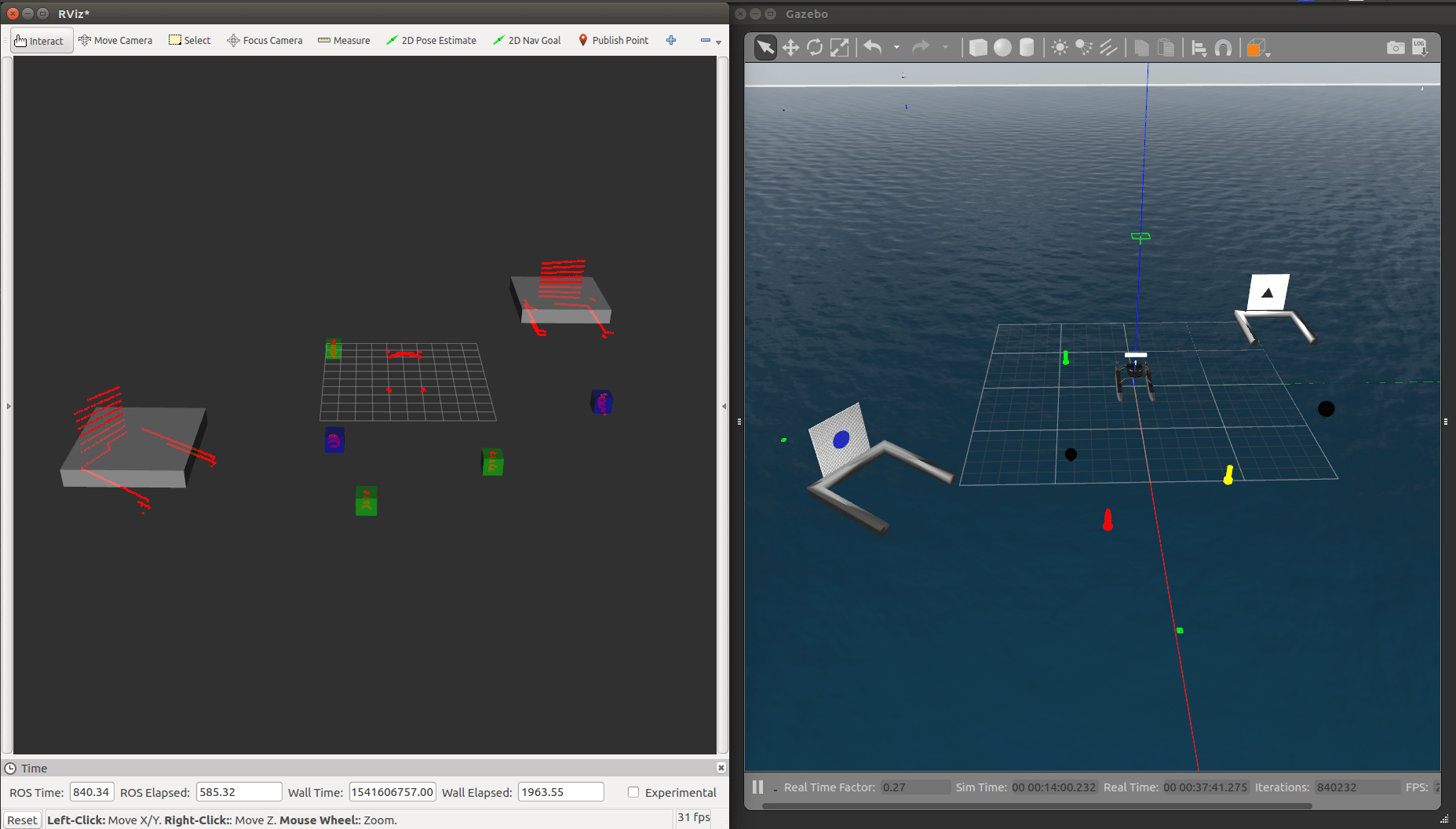}
        \caption{We test the object classification in marine Gazebo environment, and get an well performance}
 \label{figure:gazebo_classify}
\end{figure}

After collecting our dataset, we train them with CaffeNet and got an accuracy of 95.4\% in gazebo testing dataset Figure~\ref{figure:gazebo_classify} and 87.7\% in real world testing dataset. The decrease of accuracy in the real world testing dataset is mainly because the object is too far from the LIDAR. The obtained sparse pointcloud leads to worse performance. But overall, it is still a decent algorithm for the tasks in RobotX.

\subsection{Placard Detection}
\begin{figure}[ht]
  \centering
    \includegraphics[width=0.8\columnwidth]{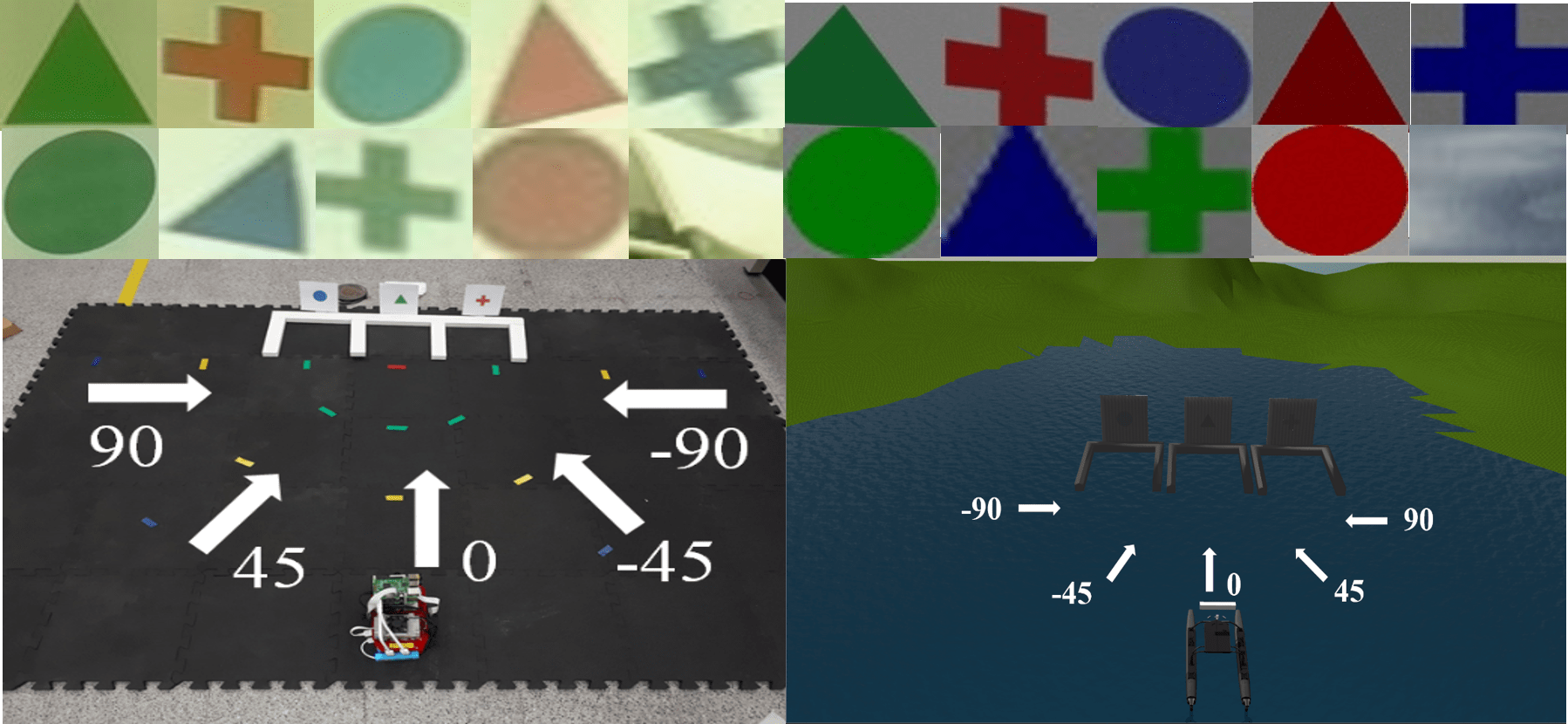}
        \caption{The training data for CNN placard classifier was collected from virtual and real environments by driving vehicle following different trajectories. The data contains data with a variety of perspective transformation, with occlusion, and illumination variance. Left: semi-realistic environment Duckietown~\cite{duckietown-icra}. Right: virtual environment Gazebo.}
 \label{figure:placard_data_collection}
\end{figure}

\begin{figure}[ht]
  \centering
    \includegraphics[width=0.8\columnwidth]{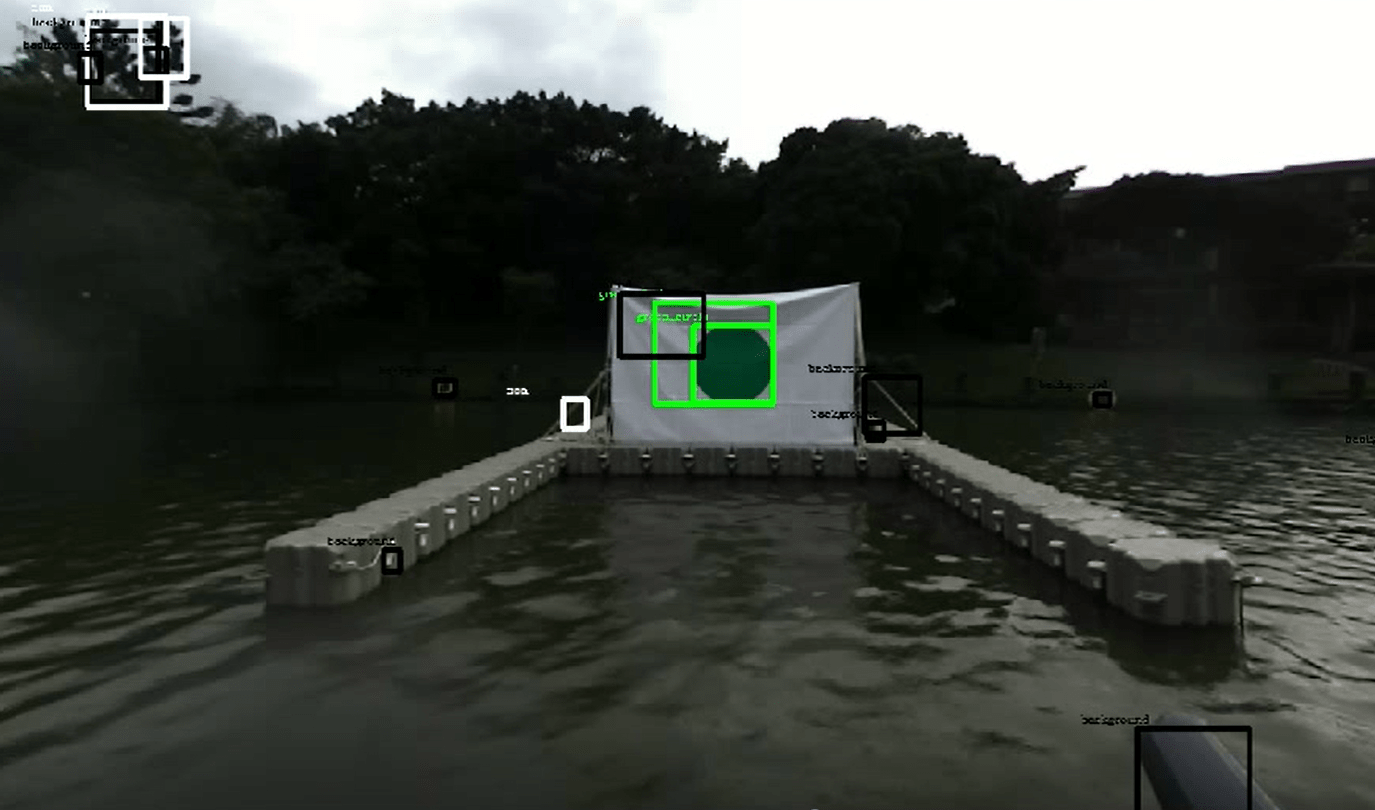}
        \caption{The CNN classifier trained from virtual and semi-realistic environments is applied to identify placard in real-world environments. Green boxes: green circle class. White boxes: predictions under threshold. Black boxes: background class.}
 \label{figure:placard_prediction}
\end{figure}

For the placard detection algorithm in the task identify the dock, most of the traditional methods based on feature matching suffered from illumination variance, perspective transformation, and occlusion. And the methods were limited to scaling up with more types of placards. To overcome those challenges, a deep CNN classifier was utilized to identify the placards. The training data was collected from virtual and semi-realistic environments which are more accessible. We collect data by driving the vehicle in different trajectories shown in Fig.~\ref{figure:placard_data_collection}, the data contained different perspective transformation, and some with occlusion. Our CNN model is based on CaffeNet~\cite{jia2014caffe} but using network surgery techniques to reconfigure it for having 10 output classess. The 10 classes represent nine different types of placards and one background. After training from virtual and semi-realistic environments, the CNN classifier is applied to perform placard identification with MSER region proposals input in real-world environment. As shown in Fig.~\ref{figure:placard_prediction}, the green, black, and white boxes represent the green circle class, background class, and predictions under threshold respectively.

\subsection{Docking Motion}
\begin{figure}[hb]
  \centering
    \includegraphics[width=0.8\columnwidth]{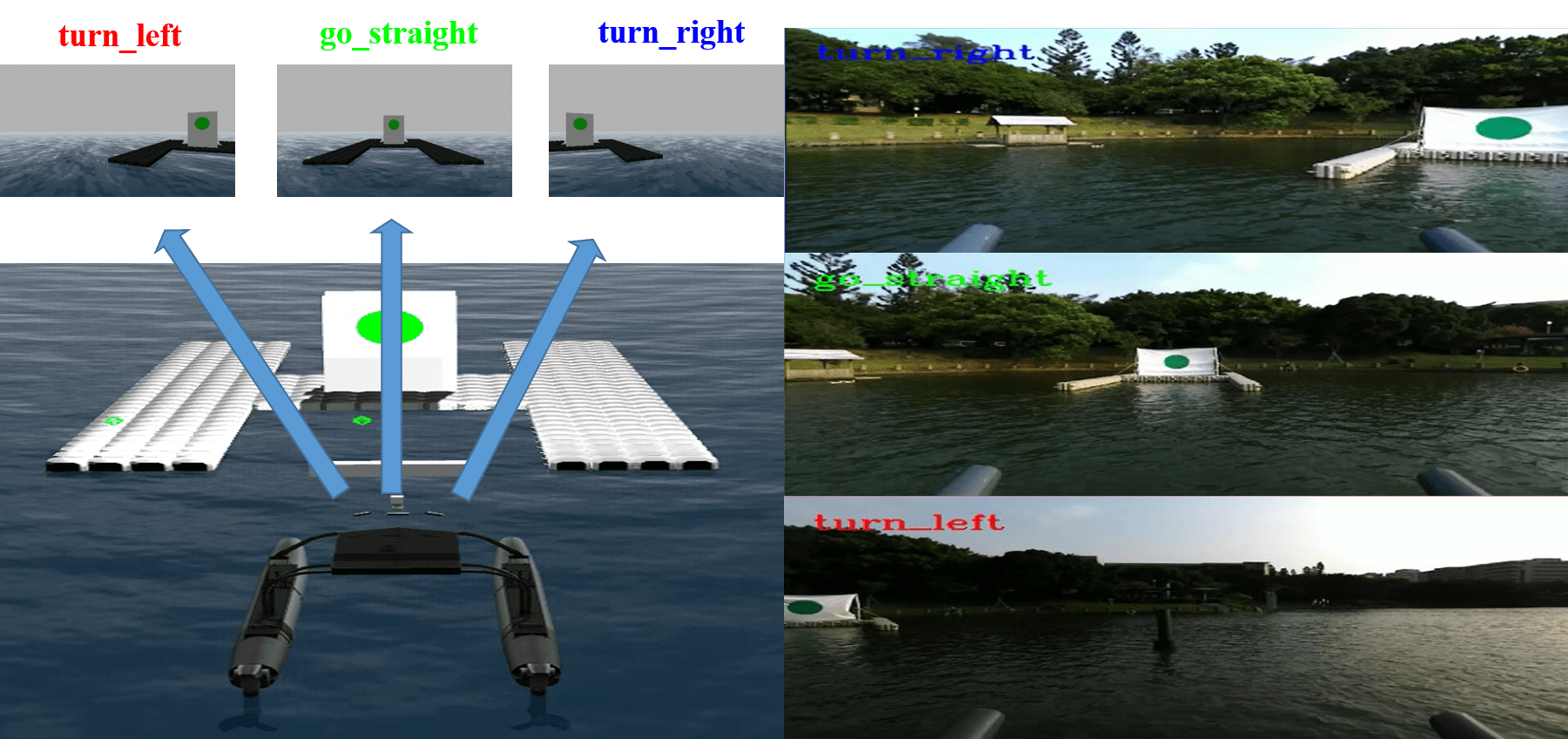}
        \caption{An end-to-end deep CNN model was used to predict motion classes for the docking process with a single RGB image as input. Left: training data collected by three cameras with different mounting angles. The datum is automatically labeled into three motion classes. Right: real-world motion class predictions with images of a single camera equipped on the ASV.}
 \label{figure:end_to_end_docking}
\end{figure}

Inspired by the previous work~\cite{chuang2018deep}, an end-to-end deep CNN model was deployed to predict three motion probabilities with single RGB image input. The imitation learning process is to gather training data by contolling a vehicle mounted with three different heading cameras torwards a docking bay. Shown in the left of Fig.~\ref{figure:end_to_end_docking}, the data collected from three cameras was automatically labelled into three classes: turn left, go straight, turn right. The probabilities of three motion classes outputs were then calculated to the motor commands of the differential-driven vessel. The right Fig.~\ref{figure:end_to_end_docking}  showed the images from a single camera mounted on ASV with different heading angles to the docking bay in real-world environments. And the images could be predictied into correct motion classes for the following docking motion control.

\subsection{Totem Circling}

\begin{figure}[H]
  \centering
    \includegraphics[width=0.8\columnwidth]{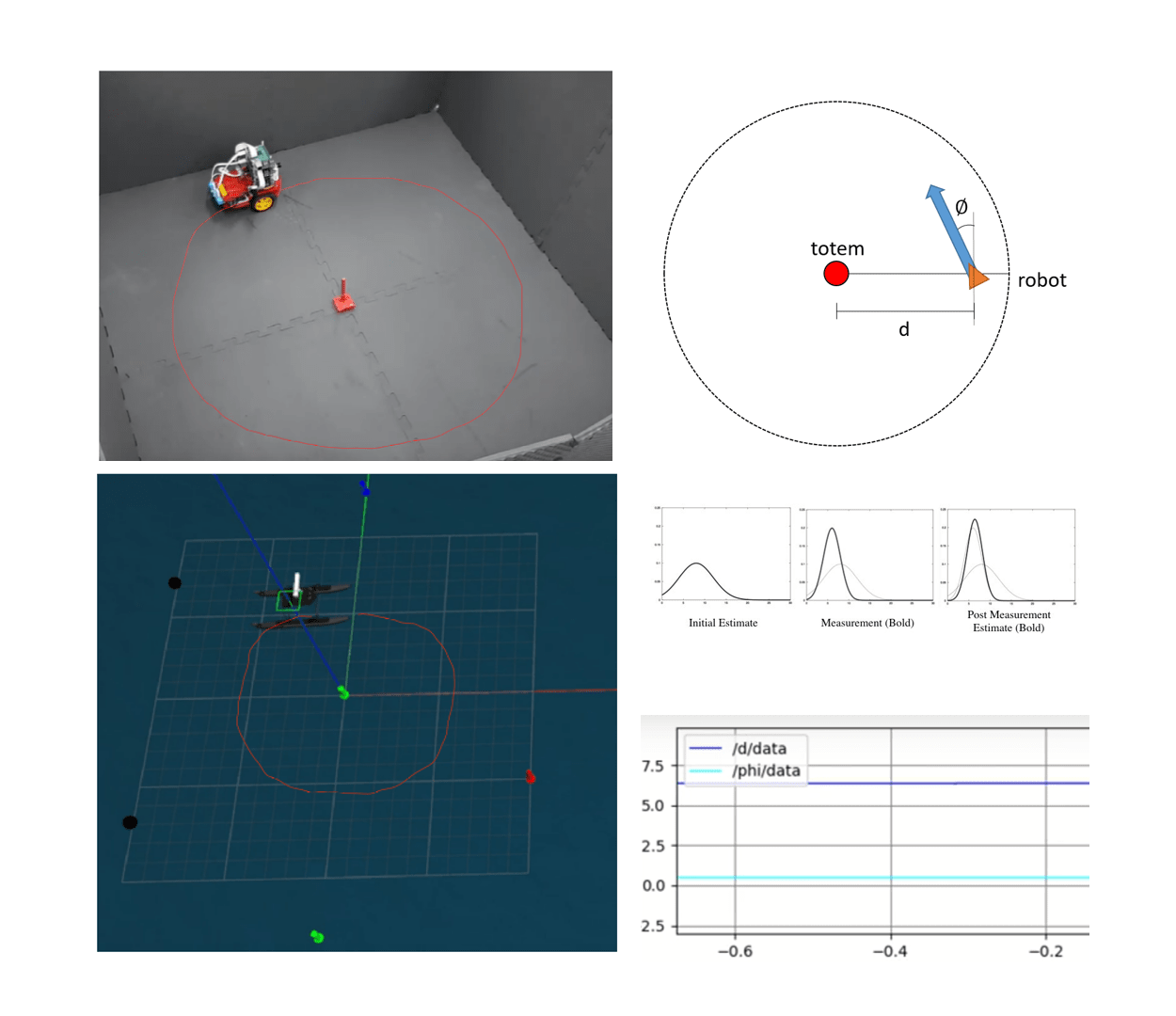}
        \caption{Robots doing totem circling in two simulation environments. The red lines show their trajectories Top Left: Duckietown. Lower left: Gazebo. Top right: the problem formulation of robot state: d and phi. Lower right: The d and phi through time for the gazbebo simulation.}
 \label{figure:circle_totem}
\end{figure}

For the totem circling task, we implemented it in the classical way. Totem detection is achieved by the object detection discussed in previous subsection therefore we could get the spatial information of the totems. Then we formulate the circling problem as figure~\ref{figure:circle_totem}. The two variables d and phi are used to describe the vehicle's position. If the vehicle is performing a circulating action, d will be the rotating radius \textit{R} and phi would be zero. Therefore, by using 2 PID controllers to control \textit{d} and \textit{phi}, the circulation motion is done. 

This algorithm is implemented in both the Duckietown and Gazebo simulations(Figurefigure~\ref{figure:circle_totem}). Both of them are robust enough to complete at least 30 rounds.

\subsection{Obstacle Avoidance}
Obstacle avoidance is one of the basic features for mobile robots and there are plenty of different algorithms to solve this problem. In this section, we are discussing three algorithms for this problem: MOOS obstacle manager, minimum angle real-time path planning and deep reinforcement learning. Different techniques may all solve the problem, but the meaning behind it could be quite different.
The first method is the obstacle avoidance feature included in MOOS. It is mainly composed of three parts. The pFeatureTracker App manages objects' convex polygon, position, and map. The pObstacleMgr App sends the OBSTACLE\_ALERT to the behavior. The BHV\_AvoidObstacle is the behavior function in MOOS-IvP Helm which maneuvers the vehicle to avoid the obstacles. In response to different situations, the user has many configuration parameters to adjust for the AvoidObstacle Behavior. This method is based on a priority policy to avoid obstacles which may be useful in many scenarios.

The second method we implemented the minimum angle real-time path planning, which is based on ~\cite{Zhuang2005real}. Basically, the algorithm forms a line segment by connecting the start point and the goal point, then checks whether this line collides with objects or not. If collision occurs, then it finds the points on the right side and left side of the obstacle as candidates. Then the turning angles of both the candidates are calculated. The candidate with a smaller angle will be picked and considered as a new starting point. Then we repeat the process of checking the new line segment that is connecting the new starting point and goal point. By doing this iteratively until the starting point is close to the goal point, we could find a path that connects the original starting point to the goal point without hitting obstacles.  

The last method we implemented for obstacle avoidance by Reinforcement learning. This is a well known problem for reinforcement learning. Work~\cite{petereinforcementobs} has been done to solve this problem using a 2D LiDAR as input with two classic reinforcement learning algorithms: Q learning and SARSA. Results are very promising in simply built simulation environments. However, with a more complex sensor input, it is common that there are thousands or even millions of robot states. This resulted in the traditional reinforcement learning almost impossible to train. Deep reinforcement learning overcomes this problem by replacing policy tables with Deep Neural Networks(DNN). We've designed our software as an AI agent(mentioned previously), so it is simple to implement it for deep reinforcement learning. 
For this task, first we set the action space as \big \langle \textit{go\_straight}, \textit{turn\_left}, \textit{turn\_right} \big \rangle. Then we downsampled the LiDAR data as the observation. Finally, we set the reward as \textit{r} a function of observation \textit{o} and action \textit{a}.The design of the reward function is simply penalize collisions and rewarding more for straight movement compared to turns for preventing the agent spinning on the same spot. By training iteratively, the DNN learns the policy. On an average of 100 episodes the agent learns to avoid obstacles in 100 steps. While this method seems workable and simple. However, some strange behaviors sometimes occur on some agents, for example, some agents only turn right to avoid obstacles. The agents only learn from the reward functions, so it makes total sense it would learn this result. Therefore it is important to carefully design the reward function.

\subsection{Underwater Manipulation}
For the underwater manipulation task we modified the inspector 2 from Seadrone Inc as our AUV. The AUV is equipped with 5 thrusters which provided 3 DoF in translation and 1 dof in orientation(yaw) movement. Onboard sensors include a pressure sensor, leak sensors, IMU and a camera with extra led lighting. We added an underwater gripper from Blue Robotics for grasping rings for the task.

Using classic image processing algorithms, we are able to detect the pose of the ring, steer the AUV to grasping point using PID control and retrieve the ring to the surface.
\begin{figure}[H]
  \centering
    \includegraphics[width=0.7\columnwidth]{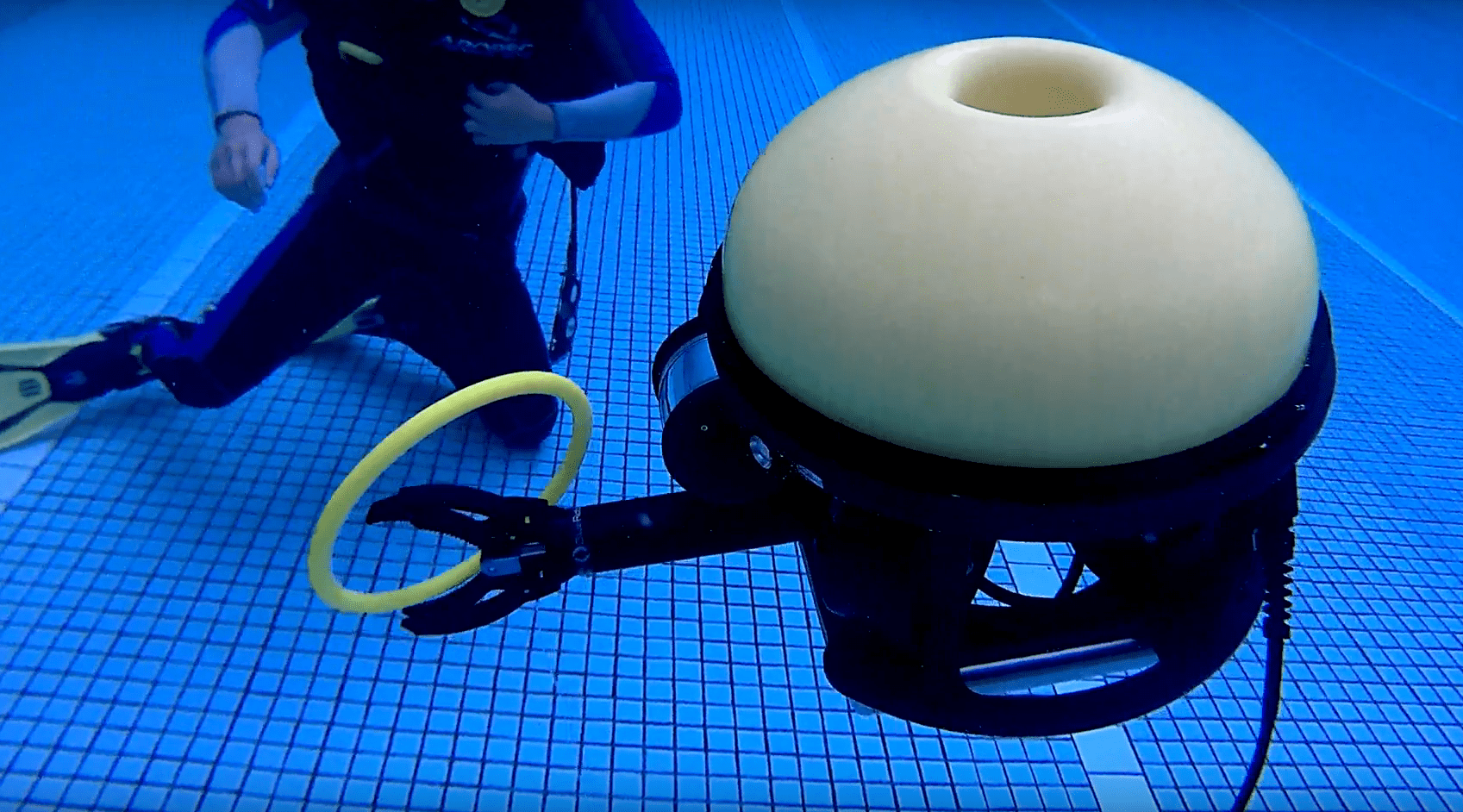}
        \caption{AUV detecting and retrieving ring autonomously}
 \label{figure:seadrone}
\end{figure}

\section{Experimental Results}

All the tests and trials of our AMS are done in the lake located in our campus. This is our first year doing this so we did a lot to get our WAM-V in the lake. We managed to build a launching platform which consists of a slope with a winching system. Neither of us had any experience so it was a trial-and-error process. We also built the equipments from totems buoys to docks by hand. 

After the WAM-V is launched into the lake, we did some test for basic functionalities(e.g. localization, motion control and perception), the problem we encountered is that we couldn't do a quantitative analysis for all the experiments related to localization. We could only do qualitative analysis and just come up to a more general conclusion. For instance, we did a trial of 10 rounds for the WAM-V performing obstacle avoidance and it completed 9 rounds. 

\begin{figure}[h]
  \centering
	\includegraphics[width=\columnwidth]{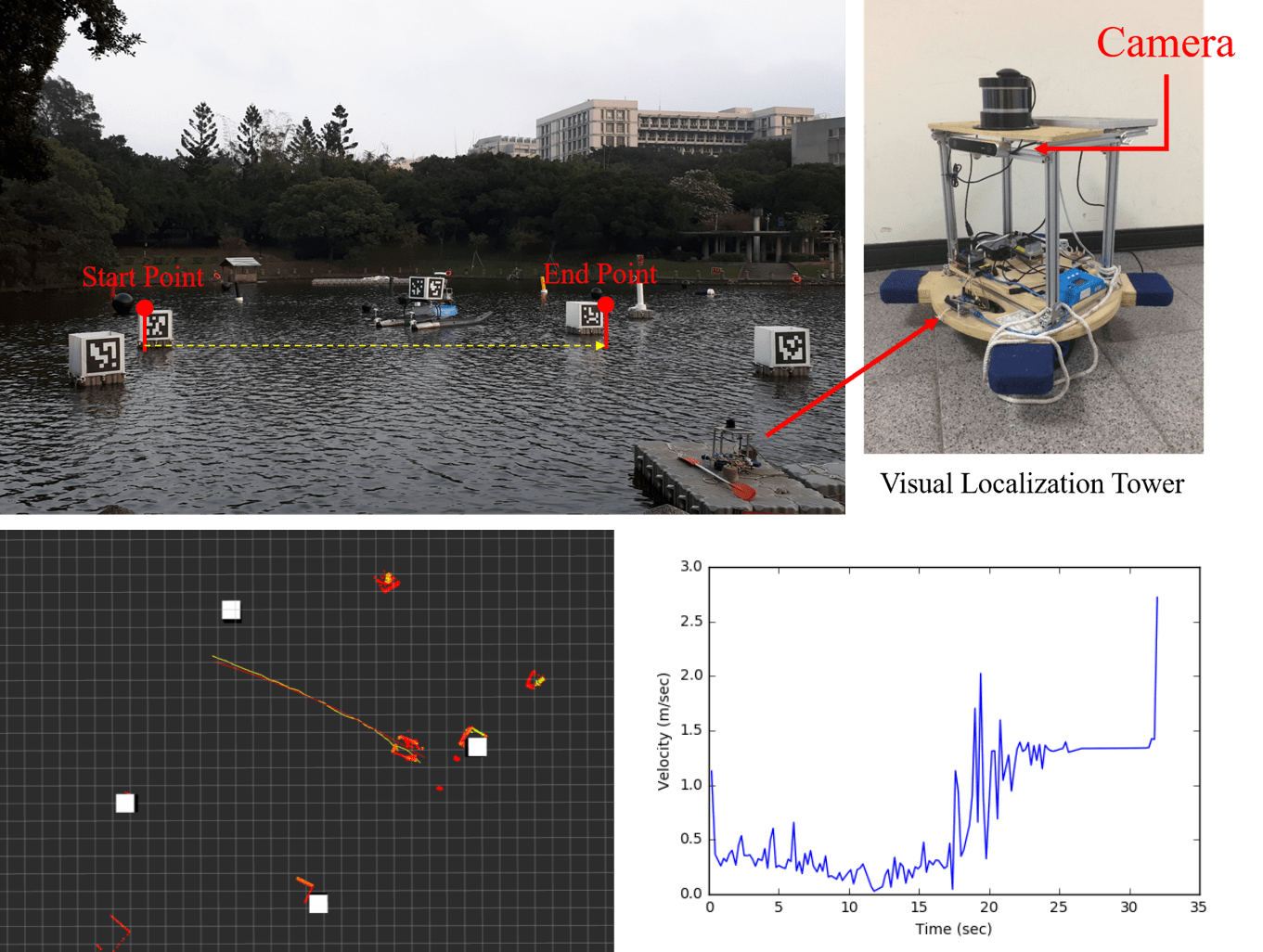}
	  \caption{The system to localize marine robotics using RGB camera with AprilTag~\cite{olson2011apriltag}. Top left: Vision-based localization system in real environment with our WAM-V. Top right: Vision-based localization tower with and RGB camera. Lower left: Trajectory in real environment using vision-based localization system.  Lower right: WAM-V velocity calculate from the vision-based localization system.}
   \label{figure:visual_localization_system}
\end{figure}

Therefore, later we decided to build a vision-based localization system Figure.~\ref{figure:visual_localization_system} by placing apriltags~\cite{olson2011apriltag} on floating pontoon cubes anchored with concrete blocks. This could provide us more quantitative results for our AMS localization, giving us more prove whether our algorithms work better. 


\section{CONCLUSIONS}

Our development of the whole system for the RobotX 2018 competition is meaningful for us being part of the unmanned autonomous vessel research. We focus not only on making our system work in the competition but also building a foundation for further researches. Our final goal is to bring a better platform and modularized software for easier development and deployment that every developer could use. This could bring the community closer together by sharing work based on the same framework and also let researchers focus on parts they care about. In our case we brought some deep learning / deep reinforcement learning algorithms and hope to bring more cutting-edge deep learning technologies to the field of marine robotics. 




\section*{ACKNOWLEDGMENT}

The research was supported by National Chiao Tung University Innovative Creative Technology (NCTU-ICT), Ministry of Science and Technology, Taiwan (grant numbers 107-2623-E-009 -005 -D and 105-2511-S-009-017-MY3), . We are also grateful for the help by Santani Teng, Robert Katzschmann, Ilenia Tinnirello, and Daniele Croce. 



\bibliographystyle{IEEEtran}
\bibliography{arg-egbib}

\begin{thebibliography}{10}
\providecommand{\url}[1]{#1}
\csname url@rmstyle\endcsname
\providecommand{\newblock}{\relax}
\providecommand{\bibinfo}[2]{#2}
\providecommand\BIBentrySTDinterwordspacing{\spaceskip=0pt\relax}
\providecommand\BIBentryALTinterwordstretchfactor{4}
\providecommand\BIBentryALTinterwordspacing{\spaceskip=\fontdimen2\font plus
\BIBentryALTinterwordstretchfactor\fontdimen3\font minus
  \fontdimen4\font\relax}
\providecommand\BIBforeignlanguage[2]{{%
\expandafter\ifx\csname l@#1\endcsname\relax
\typeout{** WARNING: IEEEtran.bst: No hyphenation pattern has been}%
\typeout{** loaded for the language `#1'. Using the pattern for}%
\typeout{** the default language instead.}%
\else
\language=\csname l@#1\endcsname
\fi
#2}}

\bibitem{aido}
\BIBentryALTinterwordspacing
The {AI} {D}riving {O}lympics. [Online]. Available:
  \url{https://nips.cc/Conferences/2018/CompetitionTrack}
\BIBentrySTDinterwordspacing

\bibitem{aido-duckietown}
\BIBentryALTinterwordspacing
The {AI} {D}riving {O}lympics. [Online]. Available:
  \url{https://AI-DO.duckietown.org}
\BIBentrySTDinterwordspacing

\bibitem{duckietown_mit}
\BIBentryALTinterwordspacing
Duckietown {MIT}. [Online]. Available: \url{http://duckietown.mit.edu/}
\BIBentrySTDinterwordspacing

\bibitem{anderson2016overview}
A.~Anderson, E.~Fischell, T.~Howe, T.~Miller, A.~Parrales-Salinas, N.~Rypkema,
  D.~Barrett, M.~Benjamin, A.~Brennen, M.~DeFillipo, \emph{et~al.}, ``An
  overview of mit-olin’s approach in the auvsi robotx competition,'' in
  \emph{Field and Service Robotics}.\hskip 1em plus 0.5em minus 0.4em\relax
  Springer, 2016, pp. 61--80.

\bibitem{benjamin2003interval}
M.~R. Benjamin, ``Interval programming: A multi-objective optimization model
  for autonomous vehicle control.'' 2003.

\bibitem{benjamin2009overview}
M.~R. Benjamin, J.~J. Leonard, H.~Schmidt, and P.~M. Newman, ``An overview of
  moos-ivp and a brief users guide to the ivp helm autonomy software,'' 2009.

\bibitem{frank2016university}
D.~Frank, A.~Gray, K.~Allen, T.~Bianchi, K.~Cohen, D.~Dugger, J.~Easterling,
  M.~Griessler, S.~Hyman, M.~Langford, \emph{et~al.}, ``University of florida:
  Team navigator ams,'' in \emph{RobotX Forum}, 2016.

\bibitem{quigley2009ros}
M.~Quigley, K.~Conley, B.~Gerkey, J.~Faust, T.~Foote, J.~Leibs, R.~Wheeler, and
  A.~Y. Ng, ``Ros: an open-source robot operating system,'' in \emph{ICRA
  workshop on open source software}, vol.~3, no. 3.2.\hskip 1em plus 0.5em
  minus 0.4em\relax Kobe, Japan, 2009, p.~5.

\bibitem{gray2016anglerfish}
A.~Gray and E.~Schwartz, ``Anglerfish: an asv controlled rov,'' in \emph{29th
  Florida Conference on Recent Advances in Robotics (FCRAR), Miami, FL}, 2016.

\bibitem{giusti2016machine}
A.~Giusti, J.~Guzzi, D.~C. Cire{\c{s}}an, F.-L. He, J.~P. Rodr{\'\i}guez,
  F.~Fontana, M.~Faessler, C.~Forster, J.~Schmidhuber, G.~Di~Caro,
  \emph{et~al.}, ``A machine learning approach to visual perception of forest
  trails for mobile robots,'' \emph{IEEE Robotics and Automation Letters},
  vol.~1, no.~2, pp. 661--667, 2016.

\bibitem{chen2015deepdriving}
C.~Chen, A.~Seff, A.~Kornhauser, and J.~Xiao, ``Deepdriving: Learning
  affordance for direct perception in autonomous driving,'' in
  \emph{Proceedings of the IEEE International Conference on Computer Vision},
  2015, pp. 2722--2730.

\bibitem{peng2015learning}
X.~Peng, B.~Sun, K.~Ali, and K.~Saenko, ``Learning deep object detectors from
  3d models,'' in \emph{Proceedings of the IEEE International Conference on
  Computer Vision}, 2015, pp. 1278--1286.

\bibitem{su2015render}
H.~Su, C.~R. Qi, Y.~Li, and L.~J. Guibas, ``Render for cnn: Viewpoint
  estimation in images using cnns trained with rendered 3d model views,'' in
  \emph{Proceedings of the IEEE International Conference on Computer Vision},
  2015, pp. 2686--2694.

\bibitem{tzeng2015towards}
E.~Tzeng, C.~Devin, J.~Hoffman, C.~Finn, X.~Peng, S.~Levine, K.~Saenko, and
  T.~Darrell, ``Towards adapting deep visuomotor representations from simulated
  to real environments,'' \emph{CoRR, abs/1511.07111}, 2015.

\bibitem{tzeng2015simultaneous}
E.~Tzeng, J.~Hoffman, T.~Darrell, and K.~Saenko, ``Simultaneous deep transfer
  across domains and tasks,'' in \emph{Proceedings of the IEEE International
  Conference on Computer Vision}, 2015, pp. 4068--4076.

\bibitem{zhang2015towards}
F.~Zhang, J.~Leitner, M.~Milford, B.~Upcroft, and P.~Corke, ``Towards
  vision-based deep reinforcement learning for robotic motion control,''
  \emph{arXiv preprint arXiv:1511.03791}, 2015.

\bibitem{zhu2017target}
Y.~Zhu, R.~Mottaghi, E.~Kolve, J.~J. Lim, A.~Gupta, L.~Fei-Fei, and A.~Farhadi,
  ``Target-driven visual navigation in indoor scenes using deep reinforcement
  learning,'' in \emph{Robotics and Automation (ICRA), 2017 IEEE International
  Conference on}.\hskip 1em plus 0.5em minus 0.4em\relax IEEE, 2017, pp.
  3357--3364.

\bibitem{rusu2016progressive}
A.~A. Rusu, N.~C. Rabinowitz, G.~Desjardins, H.~Soyer, J.~Kirkpatrick,
  K.~Kavukcuoglu, R.~Pascanu, and R.~Hadsell, ``Progressive neural networks,''
  \emph{arXiv preprint arXiv:1606.04671}, 2016.

\bibitem{koenig2004design}
N.~P. Koenig and A.~Howard, ``Design and use paradigms for gazebo, an
  open-source multi-robot simulator.'' in \emph{IROS}, vol.~4.\hskip 1em plus
  0.5em minus 0.4em\relax Citeseer, 2004, pp. 2149--2154.

\bibitem{brockman2016openai}
G.~Brockman, V.~Cheung, L.~Pettersson, J.~Schneider, J.~Schulman, J.~Tang, and
  W.~Zaremba, ``Openai gym,'' \emph{arXiv preprint arXiv:1606.01540}, 2016.

\bibitem{west2011overview}
M.~E. West, T.~R. Collins, J.~R. Bogle, A.~Melim, and M.~Novitzky, ``An
  overview of autonomous underwater vehicle systems and sensors at georgia
  tech,'' 2011.

\bibitem{demarco2011implementation}
K.~DeMarco, M.~E. West, and T.~R. Collins, ``An implementation of ros on the
  yellowfin autonomous underwater vehicle (auv),'' in \emph{OCEANS 2011}.\hskip
  1em plus 0.5em minus 0.4em\relax IEEE, 2011, pp. 1--7.

\bibitem{jia2014caffe}
Y.~Jia, E.~Shelhamer, J.~Donahue, S.~Karayev, J.~Long, R.~Girshick,
  S.~Guadarrama, and T.~Darrell, ``Caffe: Convolutional architecture for fast
  feature embedding,'' in \emph{Proceedings of the ACM International Conference
  on Multimedia}.\hskip 1em plus 0.5em minus 0.4em\relax ACM, 2014, pp.
  675--678.

\bibitem{duckietown-icra}
L.~Paull, J.~Tani, H.~Ahn, J.~Alonso-Mora, L.~Carlone, M.~Cap, Y.~F. Chen,
  C.~Choi, J.~Dusek, Y.~Fang, D.~Hoehener, S.-Y. Liu, M.~Novitzky, I.~F.
  Okuyama, J.~Pazis, G.~Rosman, V.~Varricchio, H.-C. Wang, D.~Yershov, H.~Zhao,
  M.~Benjamin, C.~Carr, M.~Zuber, S.~Karaman, E.~Frazzoli, D.~D. Vecchio,
  D.~Rus, J.~How, J.~Leonard, and A.~Censi, ``Duckietown: an {O}pen,
  {I}nexpensive and {F}lexible {P}latform for {A}utonomy {E}ducation and
  {R}esearch,'' in \emph{Robotics and Automation (ICRA), 2017 IEEE
  International Conference on}.\hskip 1em plus 0.5em minus 0.4em\relax IEEE,
  2017.

\bibitem{chuang2018deep}
T.-K. Chuang, N.-C. Lin, J.-S. Chen, C.-H. Hung, Y.-W. Huang, C.~Tengl,
  H.~Huang, L.-F. Yu, L.~Giarr{\'e}, and H.-C. Wang, ``Deep trail-following
  robotic guide dog in pedestrian environments for people who are blind and
  visually impaired-learning from virtual and real worlds,'' in \emph{2018 IEEE
  International Conference on Robotics and Automation (ICRA)}.\hskip 1em plus
  0.5em minus 0.4em\relax IEEE, 2018, pp. 1--7.

\bibitem{Zhuang2005real}
H.-Z. Zhuang, S.-X. Du, and T.-J. Wu, ``Real-time path planning for mobile
  robots,'' in \emph{2005 International Conference on Machine Learning and
  Cybernetics}, vol.~1, Aug 2005, pp. 526--531.

\bibitem{petereinforcementobs}
\BIBentryALTinterwordspacing
P.~F. Lucas~Manuelli, ``Reinforcement learning for autonomous driving obstacle
  avoidance using lidar.'' [Online]. Available:
  \url{http://www.peteflorence.com/ReinforcementLearningAutonomousDriving.pdf}
\BIBentrySTDinterwordspacing

\bibitem{olson2011apriltag}
E.~Olson, ``Apriltag: A robust and flexible visual fiducial system,'' in
  \emph{Robotics and Automation (ICRA), 2011 IEEE International Conference
  on}.\hskip 1em plus 0.5em minus 0.4em\relax IEEE, 2011, pp. 3400--3407.

\end{thebibliography}

\end{document}